\documentclass{article}
\usepackage{ijcai26}

\usepackage{times}
\usepackage{soul}
\usepackage{url}
\usepackage[hidelinks]{hyperref}
\usepackage[utf8]{inputenc}
\usepackage[small]{caption}
\usepackage{graphicx}
\usepackage{amsmath}
\usepackage{amsthm}
\usepackage{booktabs}
\usepackage{algorithm}
\usepackage{algorithmic}
\usepackage[switch]{lineno}
\usepackage{amssymb}
\usepackage[table]{xcolor}
\usepackage{multirow}
\usepackage[titletoc]{appendix}
\usepackage{booktabs}
\usepackage{bbm}
\usepackage{tabularx}

\usepackage{subcaption}

\newcommand{\CLS}{\text{CLS}}
\newcommand{\goodrow}{\rowcolor{yellow!20}\rule{0pt}{2.4ex}}

\newcommand{\smECE}{smECE} 

\usepackage{array}
\newcolumntype{L}[1]{>{\raggedright\arraybackslash}p{#1}}

\title{Calibration Attention: Learning Reliability-Aware Representations for Vision Transformers}

\author{
Wenhao Liang$^1$
\and
Lin Yue$^1$\and
Wei Emma Zhang$^1$\and
Miao Xu$^2$\\ 
Mingyu Guo$^1$\and
Olaf Maennel$^1$\and
Weitong Chen$^1$
\affiliations
$^1$Adelaide University\\
$^2$The University of Queensland
}

\begin{document}

\maketitle

\begin{abstract}
Most calibration methods operate at the logit level, implicitly assuming that miscalibration can be corrected without changing the underlying representation. We challenge this assumption and propose \textbf{Calibration Attention (CalAttn)}, a \emph{representation-aware} calibration module for vision transformers that couples instance-wise temperature scaling to transformer token geometry under a proper scoring objective. CalAttn predicts a sample-specific temperature from the \texttt{[CLS]} token and backpropagates calibration gradients into the backbone, thereby reshaping the uncertainty structure of the representation rather than post-hoc adjusting confidence. This yields \emph{token-conditioned uncertainty modulation} with negligible overhead (\(<0.1\%\) additional parameters). Across multiple datasets with ViT/DeiT/Swin backbones, CalAttn consistently improves calibration while preserving accuracy, achieving relative ECE reductions of \(3.7\%\) to \(77.7\%\) over strong baselines across diverse training objectives. Our results indicate that treating calibration as a representation-level problem is a practical and effective direction for trustworthy uncertainty estimation in transformers.
\end{abstract}

\begin{figure*}[h]
  \centering
  \includegraphics[width=0.67\linewidth]{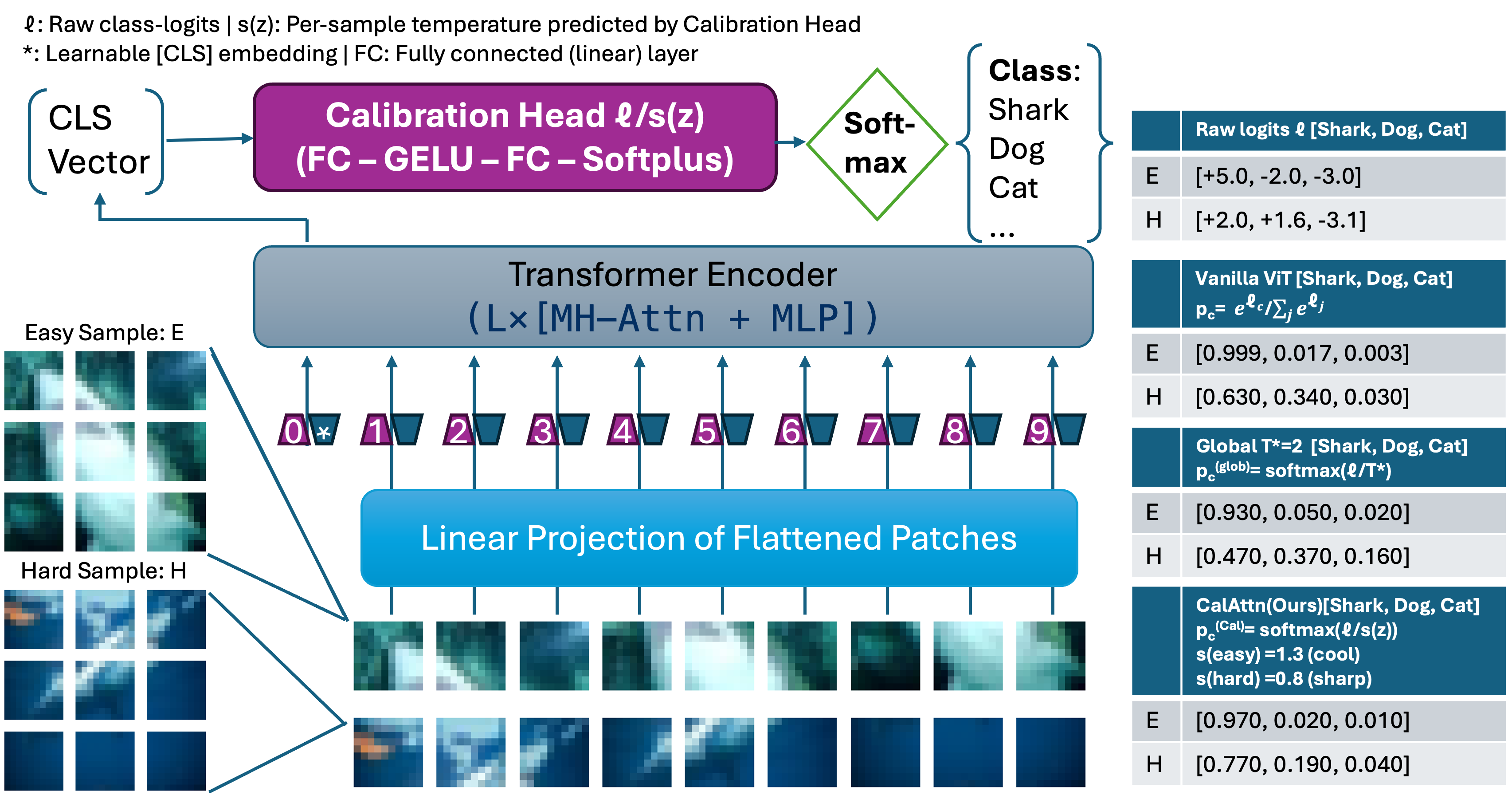}
  \caption{\textbf{Vision Transformer with Calibration Attention.}
  A lightweight calibration head reads the final \texttt{[CLS]} token and predicts a strictly positive, per-sample temperature $s(\mathbf z)$. The class logits produced by the standard classifier head are divided by this scale prior to the softmax, enabling representation-conditioned cooling or sharpening of predictive confidence. CalAttn learns reliability-aware representations by conditioning confidence on the \texttt{[CLS]} embedding, replacing global post-hoc temperature scaling.}
  \label{fig:calattn_architect}
\end{figure*}

\section{Introduction}
\label{sec:intro}

A probabilistic classifier is \emph{well calibrated} when its predicted confidence matches empirical correctness: among predictions assigned confidence $p$, roughly a $p$ fraction are correct. Calibration is therefore central to decision-making under risk and is consequential in safety-critical settings such as medical diagnosis, autonomous driving, and finance \cite{mehrtash2020confidence,feng2021review,wenhao2024enhancing}.

\paragraph{Limitations of logit-level calibration.}
Despite strong accuracy, modern deep networks---including Vision Transformers (ViTs)---are often miscalibrated \cite{guo2017calibration,minderer2021revisiting,ovadia2019can}. The canonical remedy, temperature scaling (TS), fits a single scalar temperature on a validation set and rescales logits at test time. While simple and effective in-distribution, TS implicitly assumes that miscalibration is (i) \emph{homogeneous} across inputs and (ii) \emph{expressible} as a logit-only correction. In practice, predictive uncertainty is heterogeneous and distribution-dependent: different samples exhibit different ambiguity, and shifts in data distribution can substantially change uncertainty patterns \cite{ovadia2019can}. Consequently, a global temperature can over-dampen genuinely confident predictions while under-correcting ambiguous ones, producing brittle calibration under distribution shift even when accuracy remains high. Training-time approaches (e.g., label smoothing, focal-style objectives, and differentiable calibration surrogates) mitigate miscalibration but typically rely on static hyperparameters and do not explicitly condition uncertainty on the learned representation \cite{mukhoti2020calibrating,muller2019does,tao2023dual,liang2024calibrating}.

\paragraph{Representation geometry as a source of uncertainty.}
We argue that, in transformers, miscalibration is not solely an artifact of the output layer but is tied to the geometry of the learned representation. ViTs aggregate global evidence into the \texttt{[CLS]} token, which summarizes patch information and encodes signals related to class separability, sample ambiguity, and distributional irregularities. These signals are largely invisible to post-hoc logit transformations. This motivates the view that \emph{calibration should be conditioned on internal token geometry}, rather than treated purely as a logit correction.

\paragraph{Representation-aware calibration.}
Motivated by this perspective, we introduce \emph{representation-aware calibration}: an end-to-end paradigm in which uncertainty modulation is modeled as a function of internal token representations. Concretely, we propose \textbf{Calibration Attention (CalAttn)}, a lightweight two-layer MLP that predicts an instance-specific temperature from the \texttt{[CLS]} embedding and rescales logits accordingly:
\[
T(x) = g_\phi(\mathrm{CLS}_\theta(x)),
\]
where $g_\phi$ is trained jointly with backbone parameters $\theta$. This design couples calibration with representation learning: calibration gradients propagate through the \texttt{[CLS]} channel, enabling the backbone to adjust its uncertainty geometry during training. In contrast to adaptive temperature scaling variants that regress temperatures from logits or apply post-hoc corrections on frozen features, CalAttn explicitly links token learning dynamics with uncertainty estimation in ViTs.

\paragraph{Contributions.}
Our contributions are:
\begin{itemize}
    \item \textbf{Paradigm.} We formalize \textbf{representation-aware calibration} for transformers, where uncertainty modulation is conditioned on internal token geometry rather than applied as a purely logit-level post-hoc transform.
    \item \textbf{Method.} We propose \textbf{Calibration Attention (CalAttn)}, a minimal module that predicts sample-specific temperatures from \texttt{[CLS]} and injects calibration gradients into representation via a proper scoring objective.
    \item \textbf{Evidence.} Across MNIST, CIFAR-10/100, Tiny-ImageNet, and ImageNet-1K, and across ViT/DeiT/Swin backbones, CalAttn consistently improves calibration while preserving accuracy and remains complementary to standard post-hoc methods.
\end{itemize}

\section{Method}
\label{sec:method}


We aim to replace post-hoc \emph{global} temperature scaling with an \emph{instance-wise, representation-conditioned} temperature that is learned jointly with the backbone. Given an input image, a ViT produces a final-layer \texttt{[CLS]} embedding that summarizes global evidence. CalAttn attaches a lightweight temperature head to this embedding, predicts a strictly positive scale for each sample, rescales logits before the softmax, and trains end-to-end with a calibration-aware proper scoring objective. This coupling ensures that calibration gradients propagate through the \texttt{[CLS]} pathway and can reshape the representation during training, rather than only correcting confidence after the fact. Figure~\ref{fig:calattn_architect} provides an overview; notation is summarized in Appendix~\ref{app:notations}.




\subsection{Preliminaries}
\label{sec:prelim}

\paragraph{ViT predictions.}
Given an input image $\mathbf x\in\mathbb{R}^{3\times H\times W}$, a Vision Transformer prepends a learned \texttt{[CLS]} token and, after $L$ encoder blocks, outputs the final \texttt{[CLS]} embedding $\mathbf z_{\CLS}\in\mathbb{R}^{d}$. A linear classifier head produces logits
\[
\boldsymbol\ell = \mathbf W_{\text{cls}} \mathbf z_{\CLS} + b_{\text{cls}} \in \mathbb{R}^{C},
\]
which define probabilities via the temperature-scaled softmax
\[
\sigma_i(\boldsymbol\ell, T) = \frac{\exp(\ell_i/T)}{\sum_{j=1}^{C}\exp(\ell_j/T)}.
\]

\paragraph{Global temperature scaling.}
Temperature scaling (TS) fits a single scalar $T^\star$ on a validation set and applies it at test time \cite{guo2017calibration}. This procedure is decoupled from representation learning and cannot condition confidence adjustment on sample-specific uncertainty. Under heterogeneous uncertainty or distribution shift, a single global temperature may under-correct ambiguous samples while over-smoothing confident ones.

\subsection{Calibration Attention}
\label{sec:calattn}

\paragraph{Instance-wise, representation-conditioned temperature.}
Let $\mathbf z \triangleq \mathbf z_{\CLS}$ denote the final \texttt{[CLS]} embedding. CalAttn predicts each \emph{scale} (temperature) via a two-layer MLP:
\begin{equation}
\label{eq:scale}
s(\mathbf z) =
\mathrm{Softplus}\!\left(\mathbf w_2^\top \mathrm{GELU}(\mathbf W_1 \mathbf z) + b_2 \right) + \varepsilon,
\end{equation}
where $\mathbf W_1 \in \mathbb{R}^{h\times d}$, $\mathbf w_2 \in \mathbb{R}^{h}$, $b_2\in\mathbb{R}$, and $\varepsilon=10^{-6}$ ensures strict positivity. We use an identity-preserving initialization with $\mathbf w_2=0$ and $b_2=\ln(e^{1}-1)$ such that $s(\mathbf z)\approx 1$ at the start of training, making CalAttn initially equivalent to the baseline model. The calibrated predictive distribution is
\begin{equation}
\widehat{\boldsymbol y} =
\mathrm{softmax}\!\left(\boldsymbol\ell / s(\mathbf z)\right).
\end{equation}
Since scaling preserves logit ordering, CalAttn primarily modulates confidence rather than class ranking.

\paragraph{Calibration-aware proper scoring objective.}
We train the backbone and CalAttn jointly by minimizing a proper scoring objective consisting of cross-entropy and a Brier term:
\begin{equation}
\mathcal L = \mathcal L_{\mathrm{CE}}(\widehat{\boldsymbol y},y) +
\lambda \,\|\widehat{\boldsymbol y} - \mathbf e_y\|_2^2,
\label{eq:brier}
\end{equation}
where $\mathbf e_y$ is the one-hot label and $\lambda=0.1$ unless stated otherwise. The Brier term acts as a differentiable surrogate encouraging calibrated probabilities, and its gradients backpropagate through $s(\mathbf z)$ into the \texttt{[CLS]} representation (see Appendix~\ref{app:proof-opt-s}).

\paragraph{Gradient signal through the temperature head.}
Differentiating $\mathcal L$ with respect to the (positive) scale $s$ yields
\begin{align}
\frac{\partial \mathcal{L}}{\partial s}
&=
\underbrace{\frac{\ell_y-\sum_j \hat{y}_j \ell_j}{s^2}}_{\text{CE term}}
\;+\;
\underbrace{\lambda\frac{\partial}{\partial s}\|\widehat{\boldsymbol y}-\mathbf e_y\|_2^2}_{\text{Brier term}}.
\end{align}
Using the softmax Jacobian
$J(\widehat{\boldsymbol y})=\mathrm{diag}(\widehat{\boldsymbol y})-\widehat{\boldsymbol y}\widehat{\boldsymbol y}^\top$,
the Brier contribution can be written as
\begin{equation}
\frac{\partial}{\partial s}\|\widehat{\boldsymbol y}-\mathbf e_y\|_2^2
= -\frac{2}{s^2}(\widehat{\boldsymbol y}-\mathbf e_y)^\top J(\widehat{\boldsymbol y})\boldsymbol{\ell}.
\end{equation}




\subsection{\texttt{[CLS]} Token as an Uncertainty Carrier}
\label{sec:cls_carrier}

The \texttt{[CLS]} token aggregates global context across patches and layers. Prior work suggests that its geometry encodes signals related to separability, sample difficulty, and robustness \cite{naseer2021intriguing,zhou2022understanding,minderer2021revisiting}. CalAttn leverages the full embedding to learn a non-linear mapping from representation to uncertainty modulation, rather than relying on a single handcrafted statistic. At inference, CalAttn adds a single lightweight head: compute $s(\mathbf z)$, rescale logits, and apply softmax (Algorithm~\ref{alg:calattn}).

\begin{figure}[t]
  \centering
  \includegraphics[width=\linewidth]{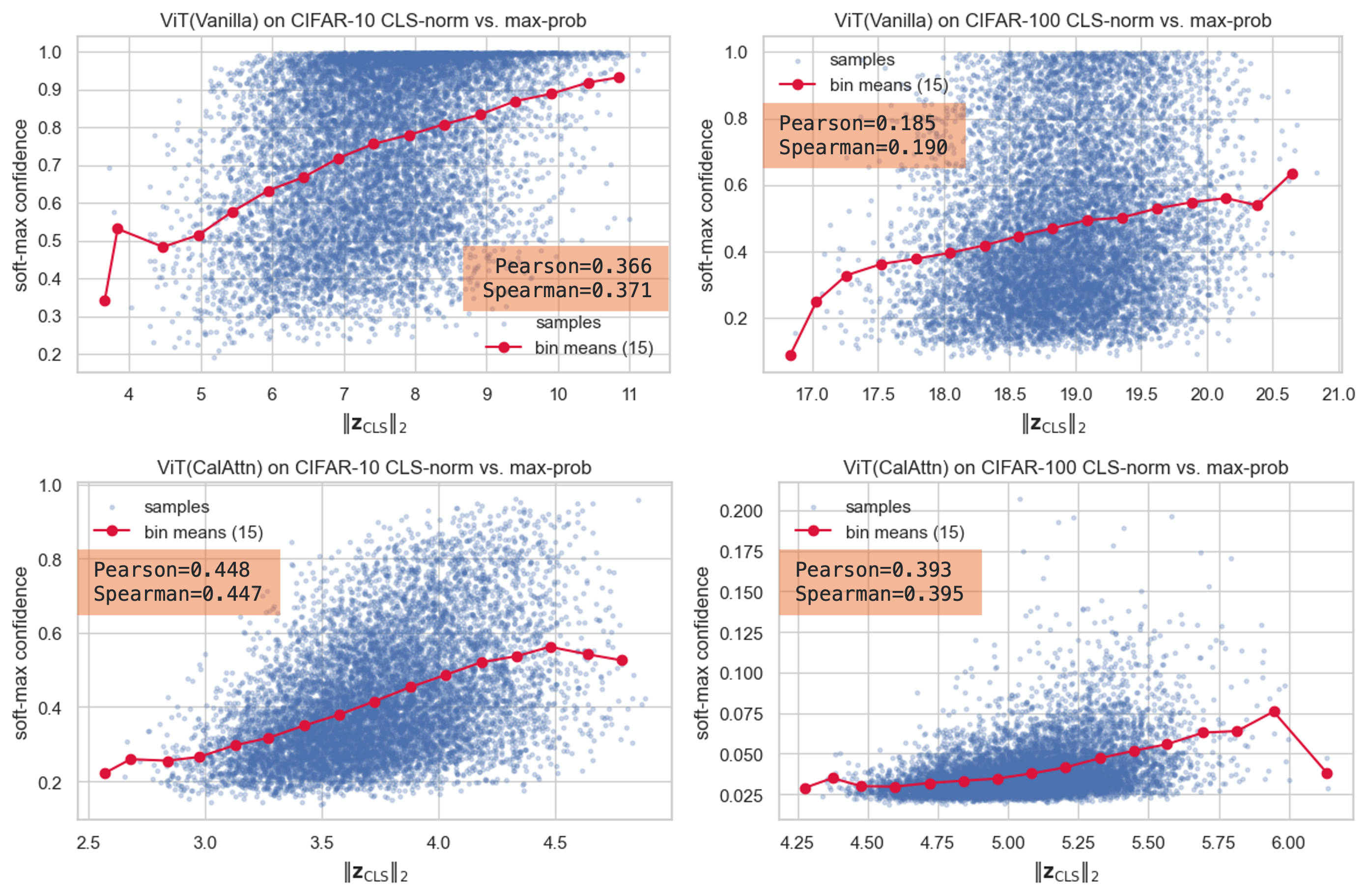}
  \caption{\texttt{[CLS]} $\ell_2$-norm versus maximum softmax probability on CIFAR-10/100. Blue dots are samples; red curves show bin means. Correlation is moderate and non-deterministic, motivating learning a representation-aware calibration function.}
  \label{fig:cls_norm_maxprob}
\end{figure}

\begin{algorithm}[tb]
\caption{Per-sample Temperature Scaling with \textbf{CalAttn}}
\label{alg:calattn}
\textbf{Input}: \texttt{[CLS]} token $\mathbf z_{\mathrm{cls}}\!\in\!\mathbb{R}^{d}$, logits $\boldsymbol\ell\!\in\!\mathbb{R}^{C}$\\
\textbf{Parameters}: $W_{1}\!\in\!\mathbb{R}^{h\times d}$, $W_{2}\!\in\!\mathbb{R}^{1\times h}$, $b_{2}=\ln(e^{1}-1)$, $\varepsilon=10^{-6}$\\
\textbf{Output}: \\
 Calibrated probabilities $\hat{\mathbf y}\!\in\!\mathbb{R}^{C}$
\begin{algorithmic}[1]
    \STATE $\mathbf h \gets \mathrm{GELU}(\mathbf W_{1}\mathbf z_{\mathrm{cls}})$ \COMMENT{first FC: $d\!\to\!h$}
    \STATE $s \gets \mathrm{Softplus}(\mathbf w_{2}^\top \mathbf h + b_{2}) + \varepsilon$ \COMMENT{$s>0$}
    \STATE $\boldsymbol\ell^{\mathrm{cal}} \gets \boldsymbol\ell / s$ \COMMENT{cool/sharpen logits}
    \STATE $\hat{\mathbf y} \gets \mathrm{softmax}(\boldsymbol\ell^{\mathrm{cal}})$
    \STATE \textbf{return} $\hat{\mathbf y}$
\end{algorithmic}
\end{algorithm}

\section{Experiments}
\label{sec:experiments}

This section provides a comprehensive evaluation of CalAttn across architectures, datasets, and calibration criteria \footnote{Code for the experiments is available at \url{https://anonymous.4open.science/r/CalibrationAttention-CalAttn--D2B8}.}.
We first report in-distribution calibration and accuracy on CIFAR-10/100, MNIST, and Tiny-ImageNet using ViT, DeiT, and Swin backbones, including results both before and after post-hoc temperature scaling.
We then analyze reliability diagrams and high-confidence errors to assess safety-critical behavior.
Next, we study robustness under distribution shift.
We further evaluate scalability on ImageNet-1K and conduct extensive ablations on calibration head design, $\lambda$ trade-off parameter, and choice of representation used to predict temperatures.
Finally, we compare CalAttn with post-hoc instance-wise methods  to quantify the benefits of learning calibration.


\subsection{Experimental Set-up}
\label{sec:exp-setup}

\paragraph{Datasets.}
We consider CIFAR-10/100 (50k/10k, $32\times32$), MNIST, Tiny-ImageNet, and ImageNet-1K, spanning increasing class counts and ambiguity regimes.

\paragraph{Architectures.}
We evaluate transformer backbones ViT$_{224}$ \cite{dosovitskiy2021vit}, DeiT-S \cite{touvron2021deit}, and Swin-S \cite{liu2021swin}. Unless explicitly stated, CalAttn augments each transformer with a single temperature head while keeping the backbone and classifier head unchanged.
(We additionally include CNN baselines where relevant for comparison to prior instance-wise calibration baselines.)

\paragraph{Training protocol.}

Unless stated otherwise, models are trained for 350 epochs with SGD (momentum 0.9, weight decay $5\times10^{-4}$) and a step learning-rate schedule: 0.1 for first 150 epochs, 0.01 for next 100 epochs, and 0.001 for final 100 epochs.
We use standard augmentation and optimization settings recommended for each backbone.
CalAttn uses $\lambda=0.1$ in Eq.~\eqref{eq:brier} throughout and we do not tune $\lambda$ per dataset.

\paragraph{Baselines.}

We re-implement representative train-time objectives including weight decay (WD) \cite{guo2017calibration}, Brier score (BS) \cite{brier1950verification}, MMCE \cite{kumar2018trainable}, label smoothing (LS) \cite{szegedy2016rethinking}, FocalLoss-53 \cite{mukhoti2020calibrating}, and Dual Focal Loss (DFL) \cite{tao2023dual}. We also report CE+$\lambda$BS (without CalAttn) to isolate the effect of using a strictly proper auxiliary term from the effect of representation-conditioned scaling.
For sample-dependent scaling baselines, we include SATS \cite{joy2023sample} and Relaxed Softmax \cite{neumann2018relaxed}.

\paragraph{Post-hoc temperature scaling (TS).}

For each trained model, TS fits a single scalar temperature $T^\star$ on a held-out validation split (5\% of the training data). Following \cite{mukhoti2020calibrating}, we select $T^\star \in \{0.1, 0.2, \ldots, 10.0\}$ by grid search to minimize validation ECE (15 equal-width bins).





\begin{table*}[ht]
\centering
\caption{\(\downarrow\) ECE (\%) before and after applying global temperature scaling (15 bins) \protect\footnotemark[2].}
\label{tab:ece_comparison}
\resizebox{\textwidth}{!}{%
\begin{tabular}{@{}cccccccccccccccc@{}}
\toprule
\multirow{2}{*}{Dataset} & \multirow{2}{*}{Model} & \multicolumn{2}{c}{\shortstack{Weight Decay \\ \cite{guo2017calibration}}} & \multicolumn{2}{c}{\shortstack{Brier Loss \\ \cite{brier1950verification}}} & \multicolumn{2}{c}{\shortstack{MMCE \\ \cite{kumar2018trainable}}} & \multicolumn{2}{c}{\shortstack{Label Smooth \\ \cite{szegedy2016rethinking}}} & \multicolumn{2}{c}{\shortstack{Focal Loss - 53 \\ \cite{mukhoti2020calibrating}}} & \multicolumn{2}{c}{\shortstack{Dual Focal \\ \cite{tao2023dual}}} & \multicolumn{2}{c}{\shortstack{CE + $\lambda$ BS\\ ($\lambda = 0.1$)} } \\ 
\cmidrule(lr){3-4} \cmidrule(lr){5-6} \cmidrule(lr){7-8} \cmidrule(lr){9-10} \cmidrule(lr){11-12} \cmidrule(lr){13-14} \cmidrule(lr){15-16}
 &  & Pre T & Post-T (T$^\star$) & Pre T & Post-T (T$^\star$) & Pre T & Post-T (T$^\star$) & Pre T & Post-T (T$^\star$) & Pre T & Post-T (T$^\star$) & Pre T & Post-T (T$^\star$) & Pre T & Post-T (T$^\star$) \\ \midrule
\multirow{9}{*}{CIFAR-100} & ResNet-50 & 18.02 & 2.60(2.20) & 5.47 & 3.71(1.10) & 15.05 & 3.41(1.90) & 6.38 & 5.12(1.10)  & 5.54 & 2.28(1.10) & 8.79 & 2.31(1.30) & 14.41 & 2.35(1.60)  \\ 
 & ResNet-110 & 19.29 & 4.75(2.30) & 6.72 & 3.59(1.20) & 18.84 & 4.52(2.30)  & 9.53 & 5.20(1.30)  & 11.02 & 3.88(1.30) & 11.65 & 3.75(1.30) & 19.36 & 3.38(1.80) \\ 
 & Wide-ResNet-26-10 & 15.17 & 2.82(2.10) & 4.07 & 3.03(1.10) & 13.57 & 3.89(2.00) & 3.29 & 3.29(1.00) & 2.62 & 2.16(1.10) & 5.30 & 2.27(1.20) & 6.63 & 5.24(1.20) \\ 
 & DenseNet-121 & 19.07 & 3.42(2.20) & 4.28 & 2.22(1.10) & 17.37 & 3.22(2.00) & 8.63 & 4.82(1.10) & 3.40 & \textbf{1.55(1.10)} & 6.69 & 1.65(1.20) & 16.10 & 2.95(1.60) \\ 
& \textbf{ViT\textsubscript{224}} &
13.11 & 3.30(1.30) &             
2.99 & 2.99(1.00) &              
11.52 & 2.34(1.30) &             
3.83 & 2.24(1.10) &              
5.33 & 3.46(1.10) &              
3.42 & 3.42(1.00) &              
6.89 & 3.49(1.50)\\
\rowcolor{yellow!20}
& \textbf{ViT\textsubscript{224}+CalAttn(Ours)} &
8.25 & 1.85(1.20) &              
\textbf{1.98} & 2.15(1.10) &              
9.04 & 2.22(1.20) &              
2.09 & 2.09(1.00) &              
\textbf{2.61} & 2.61(1.00) &              
3.72 & \textbf{1.92(1.10)} &              
4.09 & 1.49(1.10) \\

& \textbf{DeiT\textsubscript{small}} &
7.46 & 3.54(1.10) &              
2.17 & 2.17(1.00) &              
7.88 & 2.55(1.20) &              
1.75 & 1.75(1.00) &              
3.27 & 3.27(1.00) &              
3.30 & 2.19(1.10) &         
5.46 & 3.86(1.40) \\
\rowcolor{yellow!20}
& \textbf{DeiT\textsubscript{small}+CalAttn(Ours)} &
6.87 & \textbf{1.48(1.20)} &              
\textbf{1.98} & 2.19(1.10) &              
6.51 & \textbf{1.86(1.20)} &              
\textbf{1.55} & \textbf{1.55(1.00)} &              
3.15 & 3.15(1.00) &                
\textbf{1.93} & 1.93(1.00) &   
2.88 & \textbf{0.86(1.10)} \\

& \textbf{Swin\textsubscript{small}} &
3.51 & 2.46(0.90) &              
2.54 & 2.54(1.00) &              
3.53 & 2.78(0.90) &              
9.00 & 2.41(0.80) &              
10.64 & 2.51(0.80) &             
8.18 & 3.15(0.80) &              
4.59 & 2.51(1.20) \\
\goodrow
& \textbf{Swin\textsubscript{small}+CalAttn(Ours)} &
\textbf{1.93} & 1.93(1.00) &              
3.03 & \textbf{2.02(1.10)} &              
\textbf{1.92} & 1.92(1.00) &              
6.59 & 2.57(0.90) &              
6.15 & 3.17(0.90) &              
3.97 & 3.21(0.90) &              
\textbf{1.51} & 1.51(1.00) \\

\midrule
\multirow{9}{*}{CIFAR-10} & ResNet-50 & 4.23 & 1.37(2.50) & 1.83 & 0.96(1.10) & 4.67 & 1.11(2.60) & 4.23 & 1.87(0.90) & \textbf{1.46} & 1.46(1.00) & 1.32 & 1.32(1.00) & 5.71 & 1.53(1.90) \\ 
 & ResNet-110 & 4.86 & 1.94(2.60) & 2.50 & 1.36(1.20) & 5.22 & 1.24(2.80) & 3.74 & 1.22(0.90) & 1.67 & \textbf{1.10(1.10)} & 1.48 & 1.27(1.10) & 6.23 & 2.29(1.90)  \\ 
 & Wide-ResNet-26-10 & 3.24 & 1.86(2.20) & 1.17 & 1.17(1.00) & 3.60 & 1.26(2.20) & 4.66 & 1.27(0.80)  & 1.83 & 1.17(0.90) & 3.35 & \textbf{0.94(0.80)} & 3.49 & 1.15(1.50) \\ 
 & DenseNet-121 & 4.70 & 1.54(2.40) & 1.24 & 1.24(1.00) & 4.97 & 1.71(2.40) & 4.05 & 1.01(0.90) & 1.73 & 1.22(0.90) & \textbf{0.70} & 1.31(0.90) & 5.16 & 1.83(1.60) \\ 
& \textbf{ViT\textsubscript{224}} & 7.39 & 1.68(1.30) & 1.87 & 1.87(1.00) & 3.47 & 1.06(1.10) & \textbf{1.79} & 1.79(1.00) & 8.70 & 2.11(0.70) & 3.71 & 2.15(0.90) & 3.98 & 3.41(1.40) \\
\rowcolor{yellow!20}
& \textbf{ViT\textsubscript{224}+CalAttn(Ours)} & 4.76 & \textbf{1.38(1.20)} & 2.62 & 1.77(1.10) & 4.36 & 1.64(1.20) & 2.58 & 1.55(0.90) & 10.04 & 2.06(0.70) & 4.57 & 2.86(0.90) & 1.56 & 1.21(1.10) \\
& \textbf{DeiT\textsubscript{small}} & 4.42 & 1.52(1.20) & 2.02 & 2.02(1.00) & 6.01 & 1.78(1.20) & 1.84 & 1.84(1.00) & 10.65 & 2.79(0.70) & 5.11 & 2.79(0.80) & 4.25 & 2.95(1.30) \\
\rowcolor{yellow!20}
& \textbf{DeiT\textsubscript{small}+CalAttn(Ours)} & 4.10 & 1.67(1.10) & 1.98 & 1.32(1.10) & 3.57 & 1.32(1.10) & 2.41 & 2.41(1.00) & 9.91 & 1.94(0.70) & 3.87 & 2.84(0.90) & 1.88 & 1.22(1.10) \\
& \textbf{Swin\textsubscript{small}} & 1.95 & 1.91(0.90) & 1.12 & 1.12(1.00) & 1.73 & 2.31(0.90) & 6.71 & 1.45(0.80) & 15.70 & 2.32(0.60) & 8.54 & 2.40(0.70) & 2.01 & 1.47(1.20) \\
\rowcolor{yellow!20}
& \textbf{Swin\textsubscript{small}+CalAttn(Ours)} & \textbf{1.74} & 1.74(1.00) & \textbf{1.08} & \textbf{1.08(1.00)} & \textbf{1.00} & \textbf{1.00(1.00)} & 6.09 & \textbf{1.30(0.80)} & 14.97 & 1.66(0.60) & 7.50 & 2.32(0.80) & \textbf{0.94} & \textbf{0.94(1.00)} \\

\midrule
\multirow{5}{*}{MNIST} & \textbf{ViT\textsubscript{224}} & 2.91 & 1.15(0.90) & 48.83 & 5.61(8.70) & 19.98 & 5.18(2.10) & 12.77 & 8.07(1.50) & 24.07 & 2.07(1.40) & 18.12 & 4.85(2.30) & 5.34 & 1.08(0.80) \\
\rowcolor{yellow!20}
& \textbf{ViT\textsubscript{224}+CalAttn(Ours)} & 2.73 & 0.99(0.90) & 3.13 & 0.90(0.90) & 3.89 & 1.09(0.80) & 8.29 & 0.99(0.70) & \textbf{7.10} & 1.30(0.40) & 18.37 & 1.56(0.50) & \textbf{0.74} & 0.44(0.90) \\
& \textbf{DeiT\textsubscript{small}} & 2.87 & 1.15(0.90) & 28.37 & 2.73(3.20) & 4.66 & 1.16(0.80) & 8.70 & 1.47(0.70) & 21.72 & 2.43(0.50) & 15.90 & 1.55(0.70) & 4.48 & 1.04(0.80) \\
\rowcolor{yellow!20}
& \textbf{DeiT\textsubscript{small}+CalAttn(Ours)} & 1.30 & 1.30(1.00) & 1.15 & 0.46(0.90) & 2.11 & 0.92(0.90) & 6.20 & 1.00(0.80) & 18.28 & 1.54(0.50) & 8.96 & 2.28(0.60) & 0.91 & 0.91(1.00) \\
& \textbf{Swin\textsubscript{small}} & 3.64 & 3.21(1.20) & 1.32 & 0.67(0.90) & 1.26 & 0.67(0.90) & 8.09 & 0.56(0.60) & 7.09 & 1.28(0.50) & 8.78 & 0.49(0.50) & 1.86 & 1.86(1.00) \\
\rowcolor{yellow!20}
& \textbf{Swin\textsubscript{small}+CalAttn(Ours)} & \textbf{0.66} & \textbf{0.56(1.10)} & \textbf{0.85} & \textbf{0.45(0.90)} & \textbf{1.16} & \textbf{0.51(0.70)} & \textbf{6.89} & \textbf{0.46(0.60)} & 11.75 & \textbf{0.84(0.50)}  & \textbf{5.04} & \textbf{0.47(0.60)} & 0.85 & \textbf{0.42(0.90)} \\

\midrule
\multirow{5}{*}{Tiny-ImageNet} 
& \textbf{ViT\textsubscript{224}} & 22.11 & 1.93(1.90) & 4.44 & 1.89(1.10) & 19.84 & 1.81(1.80) & 2.25 & 1.51(1.10) & 3.28 & 1.93(1.10) & 2.45 & 2.44(1.10) & 1.75 & 1.66(1.10) \\
\rowcolor{yellow!20}
& \textbf{ViT\textsubscript{224}+CalAttn(Ours)} & 4.41 & 3.03(1.10) & 9.18 & 1.78(1.30) & 4.97 & 3.07(1.20) & 2.22 & 1.48(1.10) & \textbf{1.68} & 1.14(1.30) & 13.57 & 1.34(1.40) & 1.01 & 0.78(1.10) \\

& \textbf{DeiT\textsubscript{small}} & 13.04 & 2.13(1.10) & 2.70 & 1.79(1.10) & 13.66 & 1.26(1.10) & 2.36 & 1.64(1.30) & 10.28 & 2.15(1.50) & 9.37 & 1.79(1.30) & 1.69 & 1.55(1.10)  \\ \rowcolor{yellow!20}
& 
\textbf{DeiT\textsubscript{small}+CalAttn(Ours)} & \textbf{2.37} & 1.38(1.40) & 1.09 & \textbf{0.19(1.20)} & 2.83 & \textbf{1.24(1.50)} & \textbf{1.64} & \textbf{1.20(1.10)} & 2.15 & \textbf{1.11(1.30)} & \textbf{1.79} & \textbf{0.87(1.30)} & 0.95 & 0.83(1.20) \\
& \textbf{Swin\textsubscript{small}} & 3.03 & 2.76(0.90) & 0.81 & 0.25(0.90) & 1.94 & 1.94(1.00) & 4.92 & 2.41(0.90) & 4.47 & 2.46(0.90) & 4.90 & 2.57(0.90) & 1.52 & 1.52(1.00) \\\rowcolor{yellow!20}
& \textbf{Swin\textsubscript{small}+CalAttn(Ours)} & 2.92 & \textbf{1.22(1.10)} & \textbf{0.67} & 0.64(1.10) & \textbf{1.41} & 1.41(1.00) & 2.67 & 2.67(1.00) & 2.32 & 2.32(1.00) & 2.29 & 2.29(1.00) & \textbf{0.82} & \textbf{0.24(0.90)} \\

\midrule

 \bottomrule
\end{tabular}}
\end{table*}
\footnotetext[2]{Temperature scaling fits a single scalar $T^\star$ on a held-out validation split
(5\% of training data) by grid search over $T\in\{0.1,0.2,\ldots,10.0\}$ to minimize validation ECE
(15 equal-width bins), following~\cite{mukhoti2020calibrating} and utilizing their public code~\cite{muk-github-url}. Parentheses in Post-T report $T^\star$.}


\subsection{Main Results on ECE: before and after TS}
\label{sec:main-ece}

\begin{figure*}[h]
\centering
\includegraphics[width=0.95\linewidth]{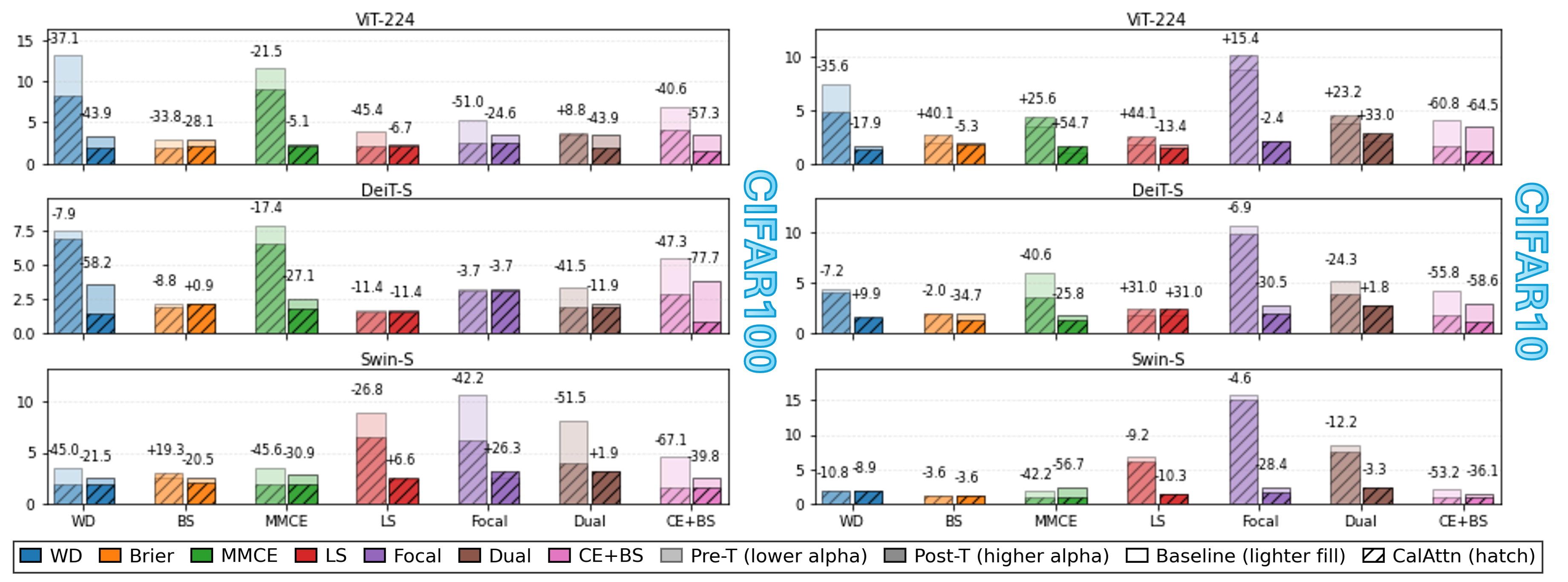}
\caption{ECE(\%) on CIFAR-10/100 across ViT-224, DeiT-S, and Swin-S with relative changes ($\Delta$ \%, “$+$” increase, “$-$” decrease).}
\label{fig:delta}
\end{figure*}

Table~\ref{tab:ece_comparison} reports ECE before and after applying temperature scaling (TS) and CalAttn yields consistent reductions(mostly negative $\Delta$\% in Fig.~\ref{fig:delta}). We highlight three findings. 

\paragraph{(1) Calibration should be considered jointly together with accuracy.}
A perfectly calibrated yet inaccurate model is undesirable. We therefore report Top-1 accuracy in Table~\ref{tab:cls_acc}, and NLL alongside calibration metrics in Table~\ref{tab:c100&10_vit_3seeds}. CalAttn improves calibration without degrading accuracy; the accuracy change is typically within a small range.


\begin{table}[t]
\centering
\caption{Top-1 classification accuracy (\%)$\uparrow$ on CIFAR-10 and CIFAR-100 validation set.
Baseline denotes cross-entropy training; ``+CalAttn'' adds the calibration head.}

\label{tab:cls_acc}
\small
\begin{tabular}{lcccc}
\toprule
\multirow{2}{*}{Backbone} & \multicolumn{2}{c}{CIFAR-10} & \multicolumn{2}{c}{CIFAR-100} \\
\cmidrule(lr){2-3}\cmidrule(lr){4-5}
 & Baseline & + CalAttn & Baseline & + CalAttn \\
\midrule
ViT\textsubscript{224}     & 66.1 & \textbf{72.8} & 59.2 &\textbf{ 63.6} \\
DeiT\textsubscript{small} & 69.4 & \textbf{76.1} & 63.9 & \textbf{67.5} \\
Swin\textsubscript{small} & 74.5 & \textbf{75.8} & 66.7 & \textbf{68.2} \\
\bottomrule
\end{tabular}
\end{table}

\paragraph{(2) Representation-conditioned scaling yields consistent gains on transformers.}
With $<0.1\%$ additional parameters (Table~\ref{tab:model_size}), CalAttn reduces ECE for ViT/DeiT/Swin across datasets. The gain often remains after TS, indicating that CalAttn improves the underlying confidence structure rather than merely shifting logits by a global scalar.

\paragraph{(3) High-confidence error analysis.}
Beyond ECE, we evaluate high-confidence false positives at threshold 0.90 (HCFP@0.90) and AUROC (Table~\ref{tab:hc_err}). CalAttn consistently reduces high-confidence mistakes, demonstrating that it mitigates the most safety-critical failure mode (over-confident errors) rather than only improving  calibration metrics.

\begin{table}[h]
\centering
\caption{\textbf{Effect of CalAttn on high-confidence errors} on CIFAR-100
after post-hoc temperature scaling.
$\Delta$ (\%) is relative change from CE+BS; negative is improvement for error/confidence,
positive is improvement for AUROC.}
\label{tab:hc_err}
\small
\setlength{\tabcolsep}{4pt}
\begin{tabular}{l l c c c}
\toprule
Backbone & Metric & CE+BS & +CalAttn & $\Delta$ (\%) \\
\midrule
ViT-224 & HCFP@0.90 $\downarrow$      & 21.99 & 17.99 & $-18.2$ \\
        & Mean Conf. $\downarrow$    & 97.47 & 92.60 & $-5.0$ \\
        & AUROC $\uparrow$         & 72.26 & 72.47 & $+0.3$ \\
\midrule
DeiT-S  & HCFP@0.90 $\downarrow$      & 11.99 & 6.99  & $-41.7$ \\
        & Mean Conf. $\downarrow$    & 96.27 & 93.50 & $-2.9$ \\
        & AUROC $\uparrow$         & 77.55 & 78.40 & $+1.1$ \\
\midrule
Swin-S  & HCFP@0.90 $\downarrow$      & 23.99 & 6.99  & $-70.9$ \\
        & Mean Conf. $\downarrow$    & 94.15 & 92.64 & $-1.6$ \\
        & AUROC $\uparrow$         & 73.71 & 81.94 & $+11.2$ \\
\bottomrule
\end{tabular}
\end{table}

\paragraph{Sensitivity to $\lambda$.}
We sweep $\lambda\in\{0.1,\dots,1.0\}$ on CIFAR-100 for CE+$\lambda$BS (Table~\ref{tab:lambda_cebs}) and CE+$\lambda$BS+CalAttn (Table~\ref{tab:lambda_full}) in Appendix~\ref{app:lambda_sensitivity}. While the vanilla loss shows mild variation, CalAttn is comparatively insensitive: $\lambda=0.1$ lies close to the best-performing region, so we fix it to avoid extra hyperparameter tuning and keep comparisons transparent.

\subsection{Fine-grained Calibration}
\label{sec:beyond-ece}

ECE with fixed-width binning can obscure localized miscalibration when confidence is skewed or when bins are unevenly populated. We therefore also report smECE (Table~\ref{tab:smece_comparison}, Fig.~\ref{fig:delta_smECE}), AdaECE (Table~\ref{tab:adaece_comparison}, Fig.~\ref{fig:delta_AdaECE}), and classwise-ECE (Table~\ref{tab:classece_comparison}) in Appendix~\ref{app:ada_class}, which address complementary issues such as unequal bin occupancy and class imbalance. CalAttn maintains consistent improvements, providing evidence that gains are not an artifact of a specific binning choice.

\subsection{Temporal behavior of CalAttn}
\label{sec:temporal}

\begin{figure*}[t]
  \centering
  \includegraphics[width=0.80\linewidth]{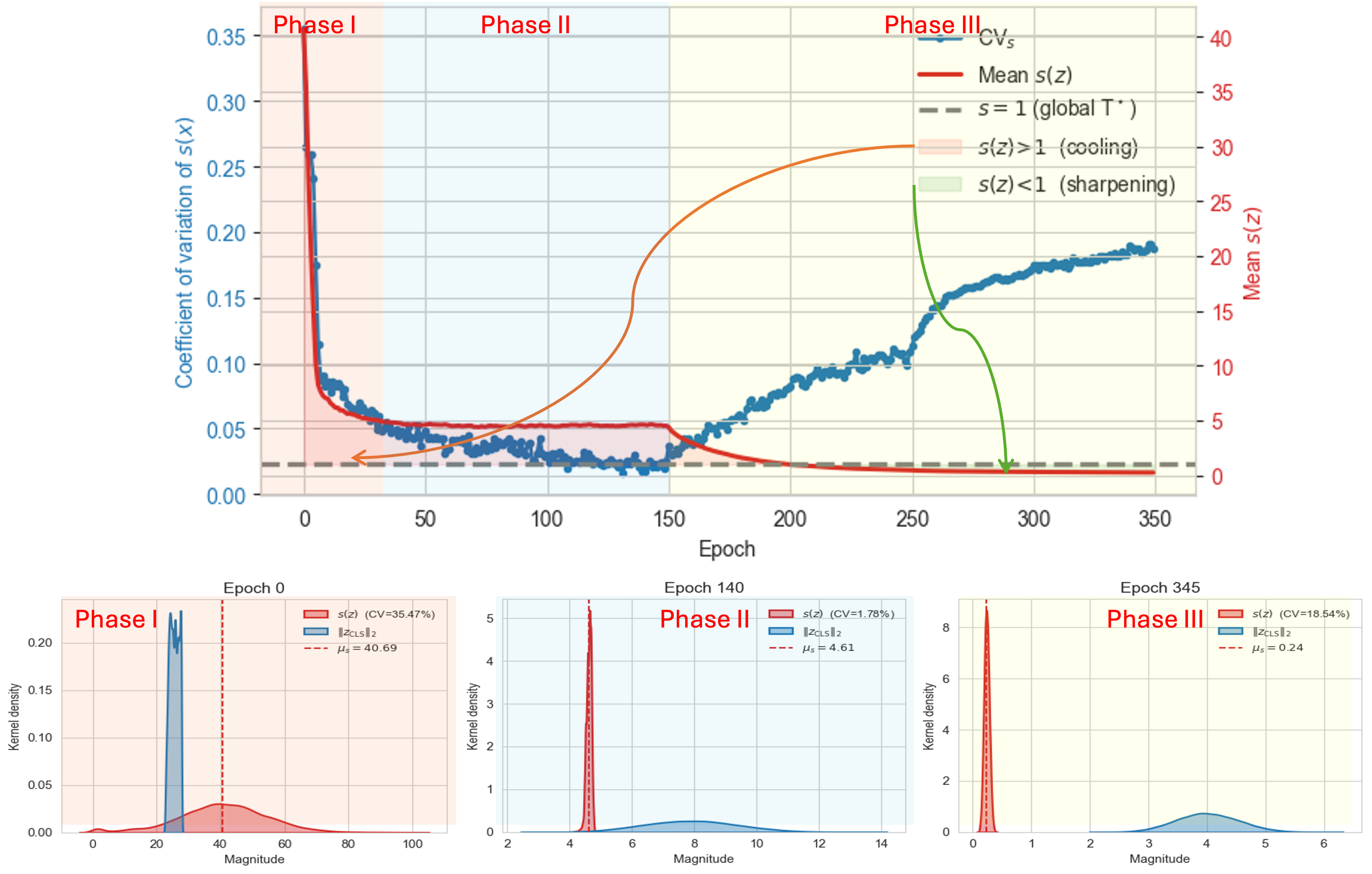}
  \caption{\textbf{Temporal dynamics of CalAttn on ViT-224 (CIFAR-10).}
\textit{Top:} mean effective temperature $\tau(x)=1/s(x)$ (red) and its coefficient of variation $\mathrm{CV}_{\tau}$ (blue) over training. The dashed line marks the optimal \emph{global} temperature $T^\star$ obtained by post-hoc temperature scaling.
\textit{Bottom:} kernel density estimates of $\tau(x)$ and $\lVert z_{\mathrm{CLS}}\rVert_2$ at three epochs.}
\label{fig:temporal_dynamics_calattn}
\end{figure*}

Figure~\ref{fig:temporal_dynamics_calattn} tracks the predicted scale
$s(x)=s(z)$ during training and reveals three qualitative regimes. CalAttn first recovers a near-global rescaling and then learns representation-conditioned deviations correlate with improved calibration.

\noindent\textbf{Phase I --- global alignment (epochs 0--30).}
The mean $s(z)$ rapidly decreases while the coefficient of variation $\mathrm{CV}_{s}$ drops, suggesting that CalAttn first behaves similarly to a near-global temperature rescaling.

\noindent\textbf{Phase II --- near-uniform modulation (epochs 30--150).}
After the first learning-rate decay, $\mathrm{CV}_{s}$ remains low and most samples receive similar scales, indicating limited instance-wise differentiation at this stage.

\noindent\textbf{Phase III --- heteroscedastic modulation (epochs 150--350).}
As training progresses, $\mathrm{CV}_{s}$ increases and CalAttn assigns more diverse scales across samples. Meanwhile, the mean scale crosses $s=1$, transitioning from cooling ($s>1$) to sharpening ($s<1$). This coincides with improved calibration, consistent with the hypothesis that representation-conditioned scaling becomes more expressive late in training.

\begin{table*}[h]
\centering
\caption{OOD robustness measured by AUROC (\%) $\uparrow$ for detecting distribution shift
from CIFAR-10 (ID) to SVHN and CIFAR-10-C.}
\label{tab:ood}
\resizebox{\textwidth}{!}{%
\begin{tabular}{l l cccccc}
\toprule
Dataset & Model &
WD~\cite{guo2017calibration} &
Brier~\cite{brier1950verification} &
MMCE~\cite{kumar2018trainable} &
LabelSmooth~\cite{szegedy2016rethinking} &
Focal~\cite{mukhoti2020calibrating} &
DualFocal~\cite{tao2023dual} \\
\midrule
\multirow{6}{*}{C10$\rightarrow$SVHN}
 & ViT-224                & 70.03 & 61.11 & 69.32 & 68.75 & 67.43 & 70.24 \\
 & ViT-224+\textbf{CalAttn} & 74.48 & 63.03 & 69.56 & 68.82 & 70.11 & \textbf{75.36} \\
 & DeiT-S               & 72.40 & 57.21 & 70.31 & 67.24 & 66.38 & 69.09 \\
 & DeiT-S+\textbf{CalAttn} & 73.72 & 63.74 & 74.53 & 74.06 & 70.43  & \textbf{76.93} \\
 & Swin-S               & 69.42 & 68.53 & 66.16 & 75.46  & 60.98 & 70.07\\
 & Swin-S+\textbf{CalAttn} & 70.43 & 52.34 & 72.45 & \textbf{78.92}& 67.67 & 72.12  \\
\midrule
\multirow{6}{*}{C10$\rightarrow$C10-C}
 & ViT-224              & 52.86 & 52.19 & 52.85 & 51.53 & 52.33 & 51.71 \\
 & ViT-224+\textbf{CalAttn} & 62.40 & \textbf{63.35} & 62.60 & 61.08 & 62.34 & 63.21 \\
 & DeiT-S               & 52.19 & 51.93 & 52.61 & 51.80 & 52.47 & 52.35 \\
 & DeiT-S+\textbf{CalAttn} & 61.50 & \textbf{63.86} & 62.34 & 59.01 & 58.27 & 60.27 \\
 & Swin-S               & 59.71 & 54.60 & 67.48 & 61.57 & 60.68 & 58.59 \\
 & Swin-S+\textbf{CalAttn} & 61.69 & 57.84 & 59.25 & 64.81 & \textbf{69.61} & 63.75 \\
\bottomrule
\end{tabular}}
\end{table*}

\subsection{Reliability Diagrams}
\label{sec:rel-plots}

\begin{figure*}[tb]
\centering
\includegraphics[width=0.85\linewidth]{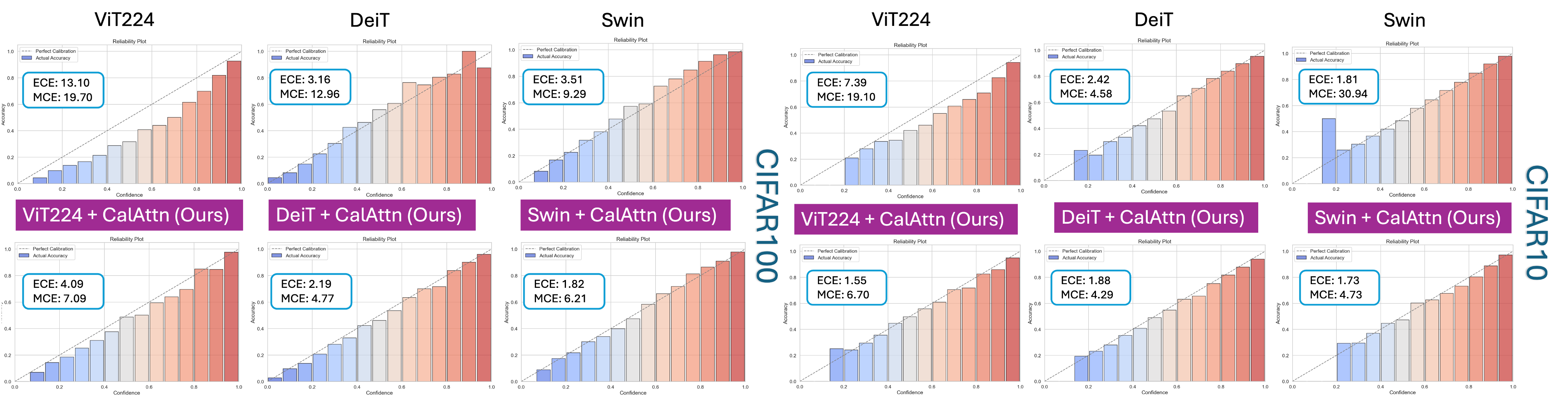}
\caption{Reliability diagrams before temperature scaling (CIFAR-10/100, 300 epochs).}
\label{fig:reliad_c10_300}
\end{figure*}

Figures~\ref{fig:reliad_c10_300} present reliability diagrams
before any post-hoc temperature scaling.
For all three backbones, the vanilla models display pronounced deviations from the
diagonal in the high-confidence regime, indicating systematic over-confidence.
In contrast, incorporating CalAttn substantially reduces these deviations, producing
curves that more closely follow the identity line.


\subsection{Robustness under Distribution Shift}
\label{sec:rob_shift}

Table~\ref{tab:ood} reports AUROC for detecting distribution shift from CIFAR-10 to SVHN and CIFAR-10-C. CalAttn improves OoD detection in most settings, with larger gains under corruption shift (CIFAR-10-C) than cross-dataset transfer (SVHN). This pattern is consistent with the intuition that corruption shifts preserve more of the ID representation geometry, whereas severe domain shifts may alter features in ways that reduce the reliability of confidence modulation.

\subsection{Ablations: inputs and head variants}
\label{sec:ablation}

We ablate the representation used to predict the temperature while keeping training settings fixed: (i) \textbf{CLS} (default), (ii) \textbf{Patch-Mean} (mean pooling of patch tokens), and (iii) \textbf{Concat} (CLS concatenated with Patch-Mean).
Table~\ref{tab:calattn_ablation} shows that Patch-Mean can improve calibration for ViT/DeiT, while CLS remains competitive for Swin, suggesting that the optimal uncertainty cue depends on how a backbone aggregates global information.
We also compare a scalar temperature head against a more expressive Dirichlet $\alpha$-head (Table~\ref{tab:dirichlet}). The scalar head achieves better calibration with fewer parameters, supporting the choice of rank-preserving temperature scaling as a stable and lightweight design.


\begin{table}[H]
\centering
\caption{\textbf{Ablation on CalAttn input features.}
Three feature choices for predicting the
per-sample temperature.}
\label{tab:calattn_ablation}
\small
\setlength{\tabcolsep}{4pt}
\begin{tabular}{l l l c c c}
\toprule
Dataset & Backbone & Feature & ECE & AdaECE & smECE \\
\midrule
\multirow{9}{*}{CIFAR-100}
 & \multirow{3}{*}{DeiT-S}
   & CLS        & 6.87 & 6.87 & 6.87 \\
 & & Patch-Mean & \textbf{4.38} & \textbf{4.41} & \textbf{4.38} \\
 & & Concat     & 7.29 & 7.29 & 7.28 \\
\cmidrule(lr){2-6}
 & \multirow{3}{*}{ViT-224}
   & CLS        & 4.76 & 8.25 & 8.24 \\
 & & Patch-Mean & \textbf{2.46} & \textbf{2.18} & \textbf{2.32} \\
 & & Concat     & 11.56 & 11.56 & 11.55 \\
\cmidrule(lr){2-6}
 & \multirow{3}{*}{Swin-S}
   & CLS        & \textbf{1.93} & \textbf{2.20} & \textbf{1.99} \\
 & & Patch-Mean & 4.43 & 4.46 & 4.40 \\
 & & Concat     & 3.02 & 3.07 & 3.03 \\
\midrule
\multirow{9}{*}{CIFAR-10}
 & \multirow{3}{*}{DeiT-S}
   & CLS        & \textbf{4.10} & \textbf{4.08} & \textbf{4.07} \\
 & & Patch-Mean & 4.89 & 4.87 & 4.85 \\
 & & Concat     & 4.20 & 4.21 & 4.21 \\
\cmidrule(lr){2-6}
 & \multirow{3}{*}{ViT-224}
   & CLS        & \textbf{4.76} & \textbf{4.47} & \textbf{4.75} \\
 & & Patch-Mean & 5.82 & 5.82 & 5.82 \\
 & & Concat     & 5.54 & 5.52 & 4.26 \\
\cmidrule(lr){2-6}
 & \multirow{3}{*}{Swin-S}
   & CLS        & \textbf{1.74} & \textbf{1.66} & \textbf{1.63} \\
 & & Patch-Mean & 1.78 & 1.90 & 1.78 \\
 & & Concat     & 1.25 & 1.39 & 1.36 \\
\bottomrule
\end{tabular}
\end{table}

\begin{table}[H]
\centering
\caption{\textbf{Scalar temperature vs.\ Dirichlet $\alpha$-head} on CIFAR-100 (ViT-224).
Lower is better for calibration metrics.}
\label{tab:dirichlet}
\small
\setlength{\tabcolsep}{4pt}
\begin{tabular}{l l c c c c}
\toprule
Head & Top-1 $\uparrow$ & ECE $\downarrow$ & AdaECE $\downarrow$ & NLL $\downarrow$ & Params $\downarrow$ \\
\midrule
Scalar (ours)        & \textbf{66.25} & \textbf{2.46} & \textbf{2.18} & \textbf{2.72} & \textbf{+0.07\%} \\
Dirichlet $\alpha$  & 64.97 & 7.33 & 7.33 & 2.89 & +0.21\% \\
\bottomrule
\end{tabular}
\end{table}



\subsection{ImageNet-1K results}
\label{sec:imagenet1k}

We evaluate CalAttn on ImageNet-1K using Swin-S (Table~\ref{tab:image1k}). CalAttn reduces ECE and AdaECE while maintaining Top-1 accuracy within a narrow range of competitive baselines. We also compare to SATS \cite{joy2023sample} under the same post-hoc TS protocol (Table~\ref{tab:sats}). While SATS performs post-hoc regression on frozen outputs, CalAttn learns representation-conditioned temperatures jointly with the backbone, yielding a different inductive bias. Appendix~\ref{app:calattn_vs_sats} provides a conceptual comparison.



\begin{table}[H]
\centering
\caption{\textbf{ImageNet-1K calibration on Swin-S} under 350 epochs.}
\label{tab:image1k}
\footnotesize
\setlength{\tabcolsep}{3pt}
\begin{tabular}{lccc}
\toprule
Method & Top-1 $\uparrow$ & ECE $\downarrow$ & AdaECE\protect\footnotemark[4] $\downarrow$ \\
\midrule
CE~\cite{guo2017calibration}           & 75.60 & 9.95 & 9.94 \\
CE+BS~\cite{brier1950verification}                & 76.80 & 4.95 & 5.42 \\
MBLS~\cite{liu2022devil}              & 77.18 & 1.95 & 1.73 \\
CSM~\cite{luo2025beyond}\protect\footnotemark[3]             & \textbf{81.08} & 1.49 & 1.86 \\
\textbf{CE+BS+CalAttn (Ours)}         & 79.68 & \textbf{1.25} & \textbf{1.43} \\
\bottomrule
\end{tabular}
\end{table}
\footnotetext[3]{CSM uses extra re-annotation and curriculum strategies.}
\footnotetext[4]{Adaptive Expected
Calibration Error (AdaECE)~\cite{ding2020revisiting} mitigates bias by adaptive-width bins with equal samples.}

\begin{table}[H]
\centering
\caption{SATS vs.\ CalAttn on CIFAR-10/100 and Tiny-ImageNet
with post-hoc temperature scaling.}
\label{tab:sats}
\footnotesize
\setlength{\tabcolsep}{3pt}
\begin{tabular}{llccc}
\toprule
Dataset & Model & Method & ECE & AdaECE\\
\midrule
CIFAR-10
 & WRN-28-10 & SATS & 1.65 & 1.61 \\
 & ResNet-50 & SATS & 1.61 & 1.53 \\
 & ViT-224   & CalAttn & 1.21 & 1.86 \\
 & DeiT-S    & CalAttn & 1.22 & 1.39 \\
 & Swin-S    & CalAttn & \textbf{0.94} & \textbf{0.88} \\
\midrule
CIFAR-100
 & WRN-28-10 & SATS & 3.67 & 3.64 \\
 & ResNet-50 & SATS & 3.30 & 3.32 \\
 & ViT-224   & CalAttn & 1.49 & \textbf{1.47} \\
 & DeiT-S    & CalAttn & \textbf{0.86} & 1.73 \\
 & Swin-S    & CalAttn & 1.51 & 1.75 \\
\midrule
Tiny-ImageNet
 & WRN-28-10 & SATS & 4.43 & 4.21 \\
 & ResNet-50 & SATS & 2.71 & 2.60 \\
 & ViT-224   & CalAttn & 0.78 & 0.75 \\
 & DeiT-S    & CalAttn & 0.83 & \textbf{0.52} \\
 & Swin-S    & CalAttn & \textbf{0.24} & 0.46 \\
\bottomrule
\end{tabular}
\end{table}

\section{Discussion}
\label{sec:discussion}

\subsection{Interpreting the gains of CalAttn}

\textbf{From latent uncertainty cues to actionable confidence modulation.}
CalAttn improves multiple calibration criteria (Tables~\ref{tab:ece_comparison} and Appendix tables), and reliability diagrams (Fig.~\ref{fig:reliad_c10_300}) show that the largest corrections occur in the high-confidence regime where over-confidence is most harmful. Operationally, CalAttn cools over-confident predictions ($s>1$) and sharpens under-confident ones ($s<1$), producing reliability curves closer to the identity line.

\noindent\textbf{A heteroscedastic view.}
The results support a heteroscedastic interpretation: the \texttt{[CLS]} embedding encodes sample-dependent uncertainty cues, and CalAttn learns a nonlinear mapping from representation to an instance-wise temperature. The temporal analysis (Fig.~\ref{fig:temporal_dynamics_calattn}) suggests that CalAttn first behaves similarly to a near-global temperature and later increases instance-wise diversity, consistent with calibration improvements that emerge as training converges.

\noindent\textbf{Adaptivity matters beyond architecture.}
After applying a single global temperature $T^\star$, the calibration gap between architectures narrows, but representation-conditioned scaling reintroduces a clear advantage by enabling sample-specific modulation. This indicates that adaptivity in confidence adjustment can be a key factor in reducing over-confidence beyond what a global post-hoc transform can achieve.

\subsection{Limitations and extensions}

CalAttn predicts a per-sample \emph{scalar} temperature, which preserves logit ordering at inference and therefore does not change top-1 decisions given fixed logits. This design is lightweight and stable, but it cannot express class-conditional or vector/matrix rescaling. Extending CalAttn to classwise temperatures, low-rank scaling, or structured Dirichlet-style parameterizations is a natural direction, trading off expressiveness against parameter count and optimization stability.

\section{Related Work}
\label{sec:related}

\textbf{Vision Transformers.}
Vision Transformers (ViT) \cite{dosovitskiy2021vit} showed that Transformer encoders can achieve competitive image classification performance. DeiT \cite{touvron2021deit} improved data efficiency via distillation, while Swin \cite{liu2021swin} introduced hierarchical windowed attention. Despite architectural refinements \cite{yuan2021tokens,chen2021visformer,dai2021coatnet}, most models retain a standard linear classification head and rely on external procedures for calibration. \textbf{Calibration in classification.}
Post-hoc calibration includes Platt scaling and temperature scaling \cite{platt1999probabilistic,guo2017calibration}, as well as more flexible variants such as vector or matrix scaling. Training-time methods include label smoothing \cite{muller2019does}, focal-style objectives \cite{mukhoti2020calibrating,tao2023dual,liang2024calibrating}, and differentiable surrogates of calibration error. Bayesian and ensemble-based approaches often improve uncertainty estimates but typically incur substantial computational overhead \cite{gal2016dropout,lakshminarayanan2017simple,ovadia2019can}. \textbf{Sample-wise temperature scaling.}
Several works predict sample-dependent temperatures using intermediate features or auxiliary predictors (e.g., Relaxed Softmax \cite{neumann2018relaxed}, Joy \emph{et al.} \cite{joy2023sample}). 

\section{Conclusion}
\label{sec:conclusion}

We study \emph{representation-conditioned} calibration, where confidence is modeled as a function of internal features. In contrast to logit-only calibrators that estimate a global or sample-dependent temperature from the output layer, CalAttn predicts a strictly positive scale from the final transformer representation and applies it as a ranking-preserving modulation. 

\newpage
\bibliographystyle{named}
\bibliography{ijcai26}

@inproceedings{guo2017calibration,
  title={On Calibration of Modern Neural Networks},
  author={Guo, Chuan and Pleiss, Geoff and Sun, Yu and Weinberger, Kilian Q},
  booktitle={International Conference on Machine Learning},
  pages={1321--1330},
  year={2017},
  organization={PMLR}
}

@inproceedings{dosovitskiy2021vit,
  title={An Image is Worth 16x16 Words: Transformers for Image Recognition at Scale},
  author={Dosovitskiy, Alexey and Beyer, Lucas and Kolesnikov, Alexander and Weissenborn, Dirk and Zhai, Xiaohua and Unterthiner, Thomas and Dehghani, Mostafa and Minderer, Matthias and Heigold, Georg and Gelly, Sylvain and Uszkoreit, Jakob and Houlsby, Neil},
  booktitle={International Conference on Learning Representations (ICLR)},
  year={2021}
}

@inproceedings{touvron2021deit,
  title={Training Data-efficient Image Transformers \& Distillation through Attention},
  author={Touvron, Hugo and Cord, Matthieu and Douze, Matthijs and Massa, Francisco and Sablayrolles, Alexandre and J{\'e}gou, Herv{\'e}},
  booktitle={International Conference on Machine Learning},
  pages={10347--10357},
  year={2021},
  organization={PMLR}
}

@Article{platt1999probabilistic,
  author        = "John Platt and others",
  title         = "Probabilistic outputs for support vector machines and comparisons to regularized likelihood methods",
  journal       = "Advances in large margin classifiers",
  volume        = "10",
  number        = "3",
  year          = "1999",
  pages         = "61--74",
  publisher     = "Cambridge, MA"
}

@Article{brier1950verification,
  author        = "Glenn W. Brier",
  title         = "Verification of forecasts expressed in terms of probability",
  journal       = "Monthly weather review",
  volume        = "78",
  number        = "1",
  year          = "1950",
  pages         = "1--3",
  publisher     = "American Meteorological Society"
}

@Article{mukhoti2020calibrating,
  author        = "Jishnu Mukhoti and Viveka Kulharia and Amartya Sanyal and Stuart Golodetz and Philip Torr and Puneet Dokania",
  title         = "Calibrating deep neural networks using focal loss",
  journal       = "Advances in Neural Information Processing Systems",
  volume        = "33",
  year          = "2020",
  pages         = "15288--15299"
}

@InProceedings{kumar2018trainable,
  author        = "Aviral Kumar and Sunita Sarawagi and Ujjwal Jain",
  title         = "Trainable calibration measures for neural networks from kernel mean embeddings",
  booktitle     = "International Conference on Machine Learning",
  year          = "2018",
  pages         = "2805--2814",
  organization  = "PMLR"
}

@Article{muller2019does,
  author        = "Rafael M{\"u}ller and Simon Kornblith and Geoffrey E. Hinton",
  title         = "When does label smoothing help?",
  journal       = "Advances in neural information processing systems",
  volume        = "32",
  year          = "2019"
}

@InProceedings{tao2023dual,
  author        = "Linwei Tao and Minjing Dong and Chang Xu",
  title         = "Dual Focal Loss for Calibration",
  booktitle     = "International Conference on Machine Learning (ICML)",
  year          = "2023"
}

@InProceedings{szegedy2016rethinking,
  author        = "Christian Szegedy and Vincent Vanhoucke and Sergey Ioffe and Jon Shlens and Zbigniew Wojna",
  title         = "Rethinking the inception architecture for computer vision",
  booktitle     = "cvpr",
  year          = "2016"
}

@article{blasiok2023smooth,
  title={Smooth ECE: Principled reliability diagrams via kernel smoothing},
  author={B{\l}asiok, Jaros{\l}aw and Nakkiran, Preetum},
  journal={arXiv preprint arXiv:2309.12236},
  year={2023}
}

@article{lakshminarayanan2017simple,
  title={Simple and scalable predictive uncertainty estimation using deep ensembles},
  author={Lakshminarayanan, Balaji and Pritzel, Alexander and Blundell, Charles},
  journal={Advances in neural information processing systems},
  volume={30},
  year={2017}
}

@article{minderer2021revisiting,
  title={Revisiting the calibration of modern neural networks},
  author={Minderer, Matthias and Djolonga, Josip and Romijnders, Rob and Hubis, Frances and Zhai, Xiaohua and Houlsby, Neil and Tran, Dustin and Lucic, Mario},
  journal={Advances in Neural Information Processing Systems},
  volume={34},
  pages={15682--15694},
  year={2021}
}

@inproceedings{wenhao2024enhancing,
  title={Enhancing Financial Market Predictions:
Causality-Driven Feature Selection},
  author={Liang, Wenhao and Li, Zhengyang and Chen, Weitong},
  booktitle={Advanced Data Mining and Applications},
  year={2024},
  publisher={Springer},
}

@misc{smce-github-url,
  author = {{Błasiok, Jarosław and Nakkiran, Preetum}},
  title = {ml-calibration},
  year = {2024},
  howpublished = {\url{https://github.com/apple/ml-calibration}},
  note = {Accessed: 2024-10-08}
}

@misc{muk-github-url,
  author = {Mukhoti, Jishnu and Kulharia, Viveka and Sanyal, Amartya and Golodetz, Stuart and Torr, Philip HS and Dokania, Puneet K},
  title = {Focal Calibration},
  year = {2020},
  howpublished = {\url{https://github.com/torrvision/focal_calibration}},
  note = {Accessed: 2024-10-08}
}

@article{feng2021review,
  title={A review and comparative study on probabilistic object detection in autonomous driving},
  author={Feng, Di and Harakeh, Ali and Waslander, Steven L and Dietmayer, Klaus},
  journal={IEEE Transactions on Intelligent Transportation Systems},
  volume={23},
  number={8},
  pages={9961--9980},
  year={2021},
  publisher={IEEE}
}

@article{mehrtash2020confidence,
  title={Confidence calibration and predictive uncertainty estimation for deep medical image segmentation},
  author={Mehrtash, Alireza and Wells, William M and Tempany, Clare M and Abolmaesumi, Purang and Kapur, Tina},
  journal={IEEE transactions on medical imaging},
  volume={39},
  number={12},
  pages={3868--3878},
  year={2020},
  publisher={IEEE}
}

@article{ovadia2019can,
  title={Can you trust your model's uncertainty? evaluating predictive uncertainty under dataset shift},
  author={Ovadia, Yaniv and Fertig, Emily and Ren, Jie and Nado, Zachary and Sculley, David and Nowozin, Sebastian and Dillon, Joshua and Lakshminarayanan, Balaji and Snoek, Jasper},
  journal={Advances in neural information processing systems},
  volume={32},
  year={2019}
}

@article{liang2024calibrating,
  title={Calibrating Deep Neural Network using Euclidean Distance},
  author={Liang, Wenhao and Dong, Chang and Zheng, Liangwei and Li, Zhengyang and Zhang, Wei and Chen, Weitong},
  journal={arXiv preprint arXiv:2410.18321},
  year={2024}
}

@inproceedings{liu2021swin,
  title={Swin transformer: Hierarchical vision transformer using shifted windows},
  author={Liu, Ze and Lin, Yutong and Cao, Yue and Hu, Han and Wei, Yixuan and Zhang, Zheng and Lin, Stephen and Guo, Baining},
  booktitle={Proceedings of the IEEE/CVF international conference on computer vision},
  pages={10012--10022},
  year={2021}
}

@inproceedings{yuan2021tokens,
  title={Tokens-to-token vit: Training vision transformers from scratch on imagenet},
  author={Yuan, Li and Chen, Yunpeng and Wang, Tao and Yu, Weihao and Shi, Yujun and Jiang, Zi-Hang and Tay, Francis EH and Feng, Jiashi and Yan, Shuicheng},
  booktitle={Proceedings of the IEEE/CVF international conference on computer vision},
  pages={558--567},
  year={2021}
}

@inproceedings{chen2021visformer,
  title={Visformer: The vision-friendly transformer},
  author={Chen, Zhengsu and Xie, Lingxi and Niu, Jianwei and Liu, Xuefeng and Wei, Longhui and Tian, Qi},
  booktitle={Proceedings of the IEEE/CVF international conference on computer vision},
  pages={589--598},
  year={2021}
}

@article{dai2021coatnet,
  title={Coatnet: Marrying convolution and attention for all data sizes},
  author={Dai, Zihang and Liu, Hanxiao and Le, Quoc V and Tan, Mingxing},
  journal={Advances in neural information processing systems},
  volume={34},
  pages={3965--3977},
  year={2021}
}

@article{naseer2021intriguing,
  title={Intriguing properties of vision transformers},
  author={Naseer, Muhammad Muzammal and Ranasinghe, Kanchana and Khan, Salman H and Hayat, Munawar and Shahbaz Khan, Fahad and Yang, Ming-Hsuan},
  journal={Advances in Neural Information Processing Systems},
  volume={34},
  pages={23296--23308},
  year={2021}
}

@inproceedings{zhou2022understanding,
  title={Understanding the robustness in vision transformers},
  author={Zhou, Daquan and Yu, Zhiding and Xie, Enze and Xiao, Chaowei and Anandkumar, Animashree and Feng, Jiashi and Alvarez, Jose M},
  booktitle={International conference on machine learning},
  pages={27378--27394},
  year={2022},
  organization={PMLR}
}

@inproceedings{gal2016dropout,
  title={Dropout as a bayesian approximation: Representing model uncertainty in deep learning},
  author={Gal, Yarin and Ghahramani, Zoubin},
  booktitle={international conference on machine learning},
  pages={1050--1059},
  year={2016},
  organization={PMLR}
}

@article{luo2025beyond,
  title={Beyond One-Hot Labels: Semantic Mixing for Model Calibration},
  author={Luo, Haoyang and Tao, Linwei and Dong, Minjing and Xu, Chang},
  journal={arXiv preprint arXiv:2504.13548},
  year={2025}
}

@inproceedings{joy2023sample,
  title={Sample-dependent adaptive temperature scaling for improved calibration},
  author={Joy, Tom and Pinto, Francesco and Lim, Ser-Nam and Torr, Philip HS and Dokania, Puneet K},
  booktitle={Proceedings of the AAAI Conference on Artificial Intelligence},
  volume={37},
  number={12},
  pages={14919--14926},
  year={2023}
}

@inproceedings{liu2022devil,
  title={The devil is in the margin: Margin-based label smoothing for network calibration},
  author={Liu, Bingyuan and Ben Ayed, Ismail and Galdran, Adrian and Dolz, Jose},
  booktitle={Proceedings of the IEEE/CVF Conference on Computer Vision and Pattern Recognition},
  pages={80--88},
  year={2022}
}

@INPROCEEDINGS {ding2020revisiting,
author = { Ding, Yukun and Liu, Jinglan and Xiong, Jinjun and Shi, Yiyu },
booktitle = { 2020 IEEE/CVF Conference on Computer Vision and Pattern Recognition Workshops (CVPRW) },
title = {{ Revisiting the Evaluation of Uncertainty Estimation and Its Application to Explore Model Complexity-Uncertainty Trade-Off }},
year = {2020},
volume = {},
ISSN = {},
pages = {22-31},
doi = {10.1109/CVPRW50498.2020.00010},
url = {https://doi.ieeecomputersociety.org/10.1109/CVPRW50498.2020.00010},
publisher = {IEEE Computer Society},
address = {Los Alamitos, CA, USA},
month =Jun}

@article{neumann2018relaxed,
  title={Relaxed softmax: Efficient confidence auto-calibration for safe pedestrian detection},
  author={Neumann, Lukas and Zisserman, Andrew and Vedaldi, Andrea},
  year={2018},
  publisher={OpenReview}
}

\clearpage
\appendix

\label{sec:appendix}
\section*{APPENDIX}

\section{Notation for Calibration Attention}
\label{app:notations}

Table~\ref{tab:notation_main} and Table~\ref{tab:notation_metrics} summarize the symbols used in the paper.
Shapes correspond to vision classification unless otherwise specified.

\begin{table}[h]
\centering
\caption{Notation for CalAttn (model and training).}
\label{tab:notation_main}
\footnotesize
\setlength{\tabcolsep}{3pt}
\renewcommand{\arraystretch}{1.05}
\begin{tabular}{p{0.20\columnwidth} p{0.52\columnwidth} p{0.22\columnwidth}}
\toprule
\textbf{Symbol} & \textbf{Description} & \textbf{Shape / Type} \\
\midrule
$B$ & batch size & scalar \\
$C$ & number of classes & scalar \\
$d$ & \texttt{[CLS]} embedding dim. & 384 / 768 \\
$H,W$ & input height / width & e.g.\ 32, 224 \\
\midrule
$\mathbf x$ & input image & $(B,3,H,W)$ \\
$\mathbf z_{\CLS}$ & final \texttt{[CLS]} embedding & $(B,d)$ \\
$W_{\text{cls}}, b_{\text{cls}}$ & classifier parameters & $(C,d),\,(C)$ \\
$\boldsymbol\ell$ & logits $W_{\text{cls}}\mathbf z_{\CLS}+b_{\text{cls}}$ & $(B,C)$ \\
$\hat{\mathbf y}$ & probs.\ $\mathrm{softmax}(\boldsymbol\ell)$ & $(B,C)$ \\
\midrule
$h$ & CalAttn hidden width & 128 \\
$W_1$ & CalAttn first-layer weight & $(h,d)$ \\
$\mathbf w_2, b_2$ & CalAttn second-layer weight/bias & $(h),\,()$ \\
$\mathbf h$ & $\mathrm{GELU}(W_1\mathbf z_{\CLS})$ & $(B,h)$ \\
$s(\mathbf z)$ & predicted scale ($>0$) & $(B,1)$ \\
$\varepsilon$ & numerical constant & $10^{-6}$ \\
$\tilde{\boldsymbol\ell}$ & scaled logits $\boldsymbol\ell/s(\mathbf z)$ & $(B,C)$ \\
$\widehat{\mathbf y}$ & calibrated probs.\ $\mathrm{softmax}(\tilde{\boldsymbol\ell})$ & $(B,C)$ \\
$y$ & ground-truth label & $\{1,\dots,C\}$ \\
$\mathbf e_y$ & one-hot label vector & $(B,C)$ \\
$\lambda$ & Brier weight & scalar \\
\bottomrule
\end{tabular}
\end{table}

\begin{table}[t]
\centering
\caption{Notation for calibration metrics.}
\label{tab:notation_metrics}
\footnotesize
\setlength{\tabcolsep}{3pt}
\renewcommand{\arraystretch}{1.05}
\begin{tabular}{p{0.20\columnwidth} p{0.52\columnwidth} p{0.22\columnwidth}}
\toprule
\textbf{Symbol} & \textbf{Description} & \textbf{Shape / Type} \\
\midrule
$N$ & number of evaluation samples & scalar \\
$M$ & number of bins & scalar \\
$\hat p_i$ & max confidence $\max_c \hat y_{ic}$ & scalar \\
$B_m$ & fixed-width ECE bin $m$ & index set \\
$B_m^{\mathrm{ad}}$ & adaptive (equal-count) bin $m$ & index set \\
$B_{m,c}$ & class-$c$ bin $m$ (ClassECE) & index set \\
$\mathrm{acc}(\cdot)$ & bin accuracy & scalar \\
$\mathrm{conf}(\cdot)$ & bin confidence & scalar \\
ECE / MCE / AdaECE / ClassECE / smECE & calibration metrics & scalar \\
\bottomrule
\end{tabular}
\end{table}

\section{Calibration Metrics}
\label{sec:metrics}

A model is perfectly calibrated if
$\Pr(y=\hat y \mid \hat p = p)=p$ for all $p\in[0,1]$.
\paragraph{ECE / MCE / AdaECE / ClassECE.}
Let $\hat p_i=\max_c \hat y_{ic}$ and partition $[0,1]$ into $M$ bins $\{B_m\}_{m=1}^M$.
Define
\begin{align}
\text{acc}(B_m) &=
\frac{1}{|B_m|} \sum_{i \in B_m} \mathbbm{1}(\hat y_i = y_i), \\
\text{conf}(B_m) &=
\frac{1}{|B_m|} \sum_{i \in B_m} \hat p_i .
\end{align}
Then
\begin{align}
\mathrm{ECE} &= \sum_{m=1}^M \frac{|B_m|}{N}
\bigl|\text{acc}(B_m)-\text{conf}(B_m)\bigr|, \\
\mathrm{MCE} &= \max_m \bigl|\text{acc}(B_m)-\text{conf}(B_m)\bigr|.
\end{align}

\textbf{AdaECE} uses \emph{adaptive} bins $\{B_m^{\mathrm{ad}}\}_{m=1}^M$ constructed such that $|B_m^{\mathrm{ad}}|\approx N/M$:
\begin{equation}
\mathrm{AdaECE} = \sum_{m=1}^M \frac{|B_m^{\mathrm{ad}}|}{N}
\bigl|\text{acc}(B_m^{\mathrm{ad}})-\text{conf}(B_m^{\mathrm{ad}})\bigr|.
\end{equation}

\textbf{ClassECE} averages class-conditional ECE over classes. For each class $c$, let
$\hat p_i^{(c)}=\hat y_{ic}$ and bin predictions for class $c$ to obtain $\{B_{m,c}\}_{m=1}^M$.
Then
\begin{equation}
\mathrm{ClassECE}=\frac{1}{C}\sum_{c=1}^C
\sum_{m=1}^M \frac{|B_{m,c}|}{N}
\bigl|\text{acc}(B_{m,c})-\text{conf}(B_{m,c})\bigr|.
\end{equation}

\paragraph{Smooth ECE (smECE).}
\cite{blasiok2023smooth} replace bins with Gaussian kernels:

\begin{align}
\widehat{\text{acc}}(\hat p_i)
&=
\frac{\sum_j K_h(\hat p_i-\hat p_j)\,\mathbbm{1}(\hat y_j=y_j)}
     {\sum_j K_h(\hat p_i-\hat p_j)}, \\[2pt]
\smECE
&=
\frac{1}{N}\sum_i
\left| \widehat{\text{acc}}(\hat p_i) - \hat p_i \right| .
\end{align}

We fix $h=0.05$.

\paragraph{Usage in this work.}
We report all metrics above and show that CalAttn consistently reduces
each, confirming improvements beyond histogram artefacts.

\section{Vision Transformer with Calibration Attention: Forward Pass}
\label{app:vit-calattn}

\paragraph{Patch embedding.}
Given an input image $\mathbf x\in\mathbb{R}^{3\times H\times W}$, a convolutional projection
with kernel and stride $P$ maps the image to a sequence of patch tokens:
\[
\mathbf X_{\text{patch}}
= \mathrm{reshape}\!\left(
  \mathrm{Conv}^{P\times P}_{3\rightarrow D}(\mathbf x),
  N, D
\right)\in\mathbb{R}^{N\times D},
\qquad
N=\frac{HW}{P^2}.
\]

\paragraph{Class token and positional encoding.}
A learned class token $\mathbf z_{\CLS}\in\mathbb{R}^{1\times D}$ is prepended and
positional embeddings are added:
\[
\mathbf z^{(0)} =
\left[\mathbf z_{\CLS};\,\mathbf X_{\text{patch}}\right]
+ \mathbf E_{\text{pos}}
\in \mathbb{R}^{(N+1)\times D}.
\]

\paragraph{Transformer encoder blocks.}
For layers $\ell=1,\dots,L$, with pre-norm residual blocks:
\[
\begin{aligned}
\mathbf u^{(\ell)} &= \mathrm{LN}(\mathbf z^{(\ell-1)}),\\
\mathbf z^{(\ell)} &= \mathbf z^{(\ell-1)} + \mathrm{MSA}(\mathbf u^{(\ell)}),\\
\mathbf v^{(\ell)} &= \mathrm{LN}(\mathbf z^{(\ell)}),\\
\mathbf z^{(\ell)} &= \mathbf z^{(\ell)} + \mathrm{MLP}(\mathbf v^{(\ell)}).
\end{aligned}
\]

\paragraph{Calibration Attention head.}
Let $\mathbf z_{\CLS}=\mathbf z^{(L)}_{0}\in\mathbb{R}^{D}$ denote the final \texttt{[CLS]} token
(the first token of $\mathbf z^{(L)}$). CalAttn predicts a strictly positive scale
\[
s(\mathbf x)
=
\mathrm{Softplus}\!\left(\mathbf w_{2}^{\top}\,\mathrm{GELU}(\mathbf W_{1}\mathbf z_{\CLS}) + b_{2}\right)
+\varepsilon,
\]
where $\mathbf W_{1}\in\mathbb{R}^{h\times D}$, $\mathbf w_{2}\in\mathbb{R}^{h}$, $b_{2}\in\mathbb{R}$,
and $\varepsilon>0$.

\paragraph{Calibrated prediction.}
A linear classifier produces logits $\boldsymbol\ell = W_{\text{cls}}\mathbf z_{\CLS}+b_{\text{cls}}\in\mathbb{R}^{C}$.
CalAttn rescales logits and outputs calibrated probabilities:
\[
\tilde{\boldsymbol\ell}=\boldsymbol\ell / s(\mathbf x),
\qquad
\widehat{\mathbf y}=\mathrm{softmax}(\tilde{\boldsymbol\ell}).
\]

\paragraph{Training objective.}
We train end-to-end with a proper scoring objective:
\[
\mathcal L
= \mathcal L_{\mathrm{CE}}(\widehat{\mathbf y},y)
+ \lambda\left\|\widehat{\mathbf y}-\mathbf e_y\right\|_2^2,
\qquad \lambda=0.1.
\]

\section{Architectural Differences}
\label{app:arch}

Table~\ref{tab:vit_vs_calattn} contrasts a vanilla ViT with ViT equipped with Calibration Attention (CalAttn).

\begin{table*}[t]
\centering
\caption{Comparison between a vanilla Vision Transformer and ViT equipped with Calibration Attention (CalAttn).}
\label{tab:vit_vs_calattn}
\small
\setlength{\tabcolsep}{5pt}
\renewcommand{\arraystretch}{1.15}
\begin{tabularx}{\linewidth}{@{}L{3.2cm}L{3.6cm}X@{}}
\toprule
\textbf{Component} & \textbf{Vanilla ViT} & \textbf{ViT + CalAttn (Ours)} \\
\midrule
Additional module
& None
& Temperature head: lightweight two-layer MLP applied to the final \texttt{[CLS]} embedding \\

Module output
& ---
& Strictly positive, instance-wise scale $s(\mathbf z_{\CLS})$, identity-initialized so that $s(\mathbf z_{\CLS})\approx 1$ at the start of training \\

Logits before softmax
& $\boldsymbol\ell$
& $\tilde{\boldsymbol\ell}=\boldsymbol\ell / s(\mathbf z_{\CLS})$, enabling per-sample cooling ($s>1$) or sharpening ($s<1$) while preserving logit ordering \\

Training objective
& Cross-entropy (CE)
& Proper scoring objective: CE $+\ \lambda$Brier (Eq.~\eqref{eq:brier}) \\

Post-hoc temperature scaling
& Often applied as a separate post-hoc stage
& Optional; CalAttn learns instance-wise scaling during training and can be combined with global TS if desired \\

Parameter overhead
& 0
& $<0.1\%$ of backbone parameters (two small fully connected layers) \\
\bottomrule
\end{tabularx}
\end{table*}

\section{Why the Learned Scale Approximates the Conditional Optimal Temperature}
\label{app:proof-opt-s}

Let $\boldsymbol{\ell}(x)\in\mathbb{R}^{C}$ denote the logits produced by a fixed
classifier for input $x$, and let $s(x)>0$ be the CalAttn scale. Define scaled
logits $\tilde{\boldsymbol{\ell}}(x)=\boldsymbol{\ell}(x)/s(x)$ and the
predictive distribution
$\hat{\mathbf y}(x,s)=\mathrm{softmax}(\tilde{\boldsymbol{\ell}}(x))$.
The per-sample training loss is
\begin{equation}
\mathcal L(x,y;s)
  = -\log \hat y_{y}(x,s)
    + \lambda\|\hat{\mathbf y}(x,s)-\mathbf e_{y}\|_2^2 .
\end{equation}

\paragraph{Proper scoring perspective.}
For a fixed input $x$, consider the conditional risk
\[
\mathcal R_x(s)
= \mathbb{E}_{y\sim p(y\mid x)}\bigl[\mathcal L(x,y;s)\bigr].
\]
Both the cross-entropy and the Brier score are strictly proper scoring rules,
and any nonnegative weighted sum of strictly proper scoring rules is strictly
proper. Therefore, viewed as a function of the predicted distribution
$\hat{\mathbf y}$, the conditional risk is uniquely minimized at the true
posterior:
\[
\arg\min_{\hat{\mathbf y}\in\Delta^{C-1}}
\mathbb{E}_{y\sim p(y\mid x)}\!\left[-\log \hat y_y + \lambda\|\hat{\mathbf y}-\mathbf e_y\|_2^2\right]
= p(y\mid x).
\]
This identifies the \emph{target} distribution that the calibration head should
produce.

\paragraph{From distributions to a scalar scale.}
CalAttn does not predict an arbitrary distribution; it predicts a \emph{scalar}
$s(x)$ that modulates sharpness while preserving the logit ordering. Hence the
best achievable predictor under CalAttn belongs to the one-parameter family
\[
\mathcal F_x \;=\;\Bigl\{\mathrm{softmax}\!\bigl(\boldsymbol{\ell}(x)/s\bigr)\;:\;s>0\Bigr\}.
\]
Define the \emph{CalAttn-optimal} conditional scale as
\[
s^\star(x)\in\arg\min_{s>0}\; \mathcal R_x(s).
\]
When the true posterior is representable by temperature scaling, i.e., when
there exists an $s>0$ such that
$p(y\mid x)=\mathrm{softmax}(\boldsymbol{\ell}(x)/s)$, any minimizer $s^\star(x)$
satisfies
\begin{equation}
\hat{\mathbf y}(x,s^\star(x)) = p(y\mid x).
\label{eq:opt_scale_match}
\end{equation}
In general, if $p(y\mid x)\notin\mathcal F_x$, then $s^\star(x)$ yields the best
approximation to $p(y\mid x)$ within $\mathcal F_x$ under the strictly proper
conditional risk.

\paragraph{Learning with a representation-conditioned head.}
In practice, $p(y\mid x)$ is unknown and $s(x)$ is parameterized by a calibration
head conditioned on the \texttt{[CLS]} representation, $s(x)=s_\phi(z(x))$.
Training minimizes the empirical risk, so the learned head approximates the
population minimizer $s^\star(x)$ induced by the current representation. This is
the sense in which CalAttn learns a representation-conditioned approximation to
the conditional optimal temperature.

\paragraph{Proposition (Conditional optimality with frozen logits).}
Fix logits $\boldsymbol{\ell}(x)$ (equivalently, freeze backbone and classifier).
Assume $p(y\mid x)$ is well-defined for each $x$ and $\lambda\ge 0$. Let
$\hat{\mathbf y}(x,s)=\mathrm{softmax}(\boldsymbol{\ell}(x)/s)$. Then any
minimizer $s^\star(x)\in\arg\min_{s>0}\mathcal R_x(s)$ produces the best
CalAttn-representable calibrated distribution for that $x$, and if
$p(y\mid x)\in\mathcal F_x$ then Eq.~\eqref{eq:opt_scale_match} holds.

\paragraph{Remark (Scope).}
The statement is conditional on fixed logits (or frozen backbone parameters).
It does not assert global optimality for joint training over the backbone and
the calibration head. Rather, it formalizes the calibration head's role:
given a representation/logit geometry, optimizing $s_\phi$ moves predictions
toward the conditional optimum achievable by temperature scaling.

\section{smECE, AdaECE and Classwise-ECE}
\label{app:ada_class}

\begin{figure*}[h]
\centering
\includegraphics[width=\linewidth]{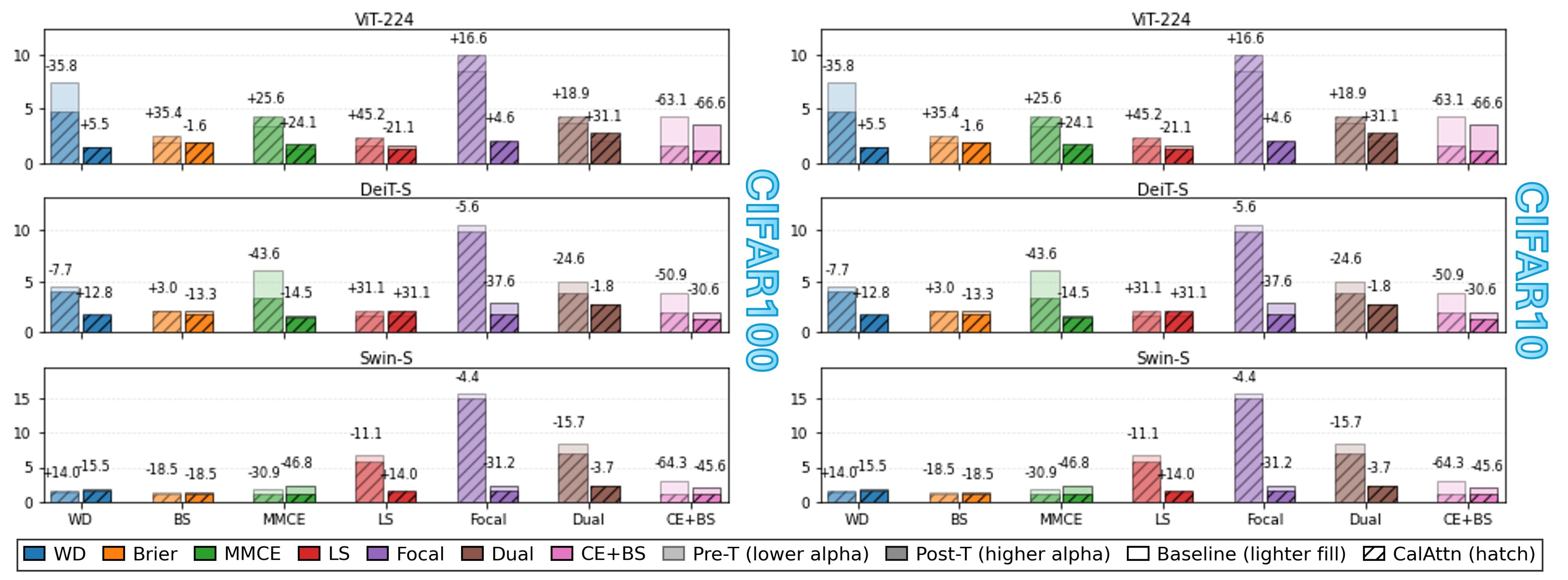}
\caption{smECE(\%) on CIFAR-10/100 across ViT-224, DeiT-S, and Swin-S with relative changes ($\Delta$ \%, “$+$” increase, “$-$” decrease).}
\label{fig:delta_smECE}
\end{figure*}

\begin{figure*}[h]
\centering
\includegraphics[width=\linewidth]{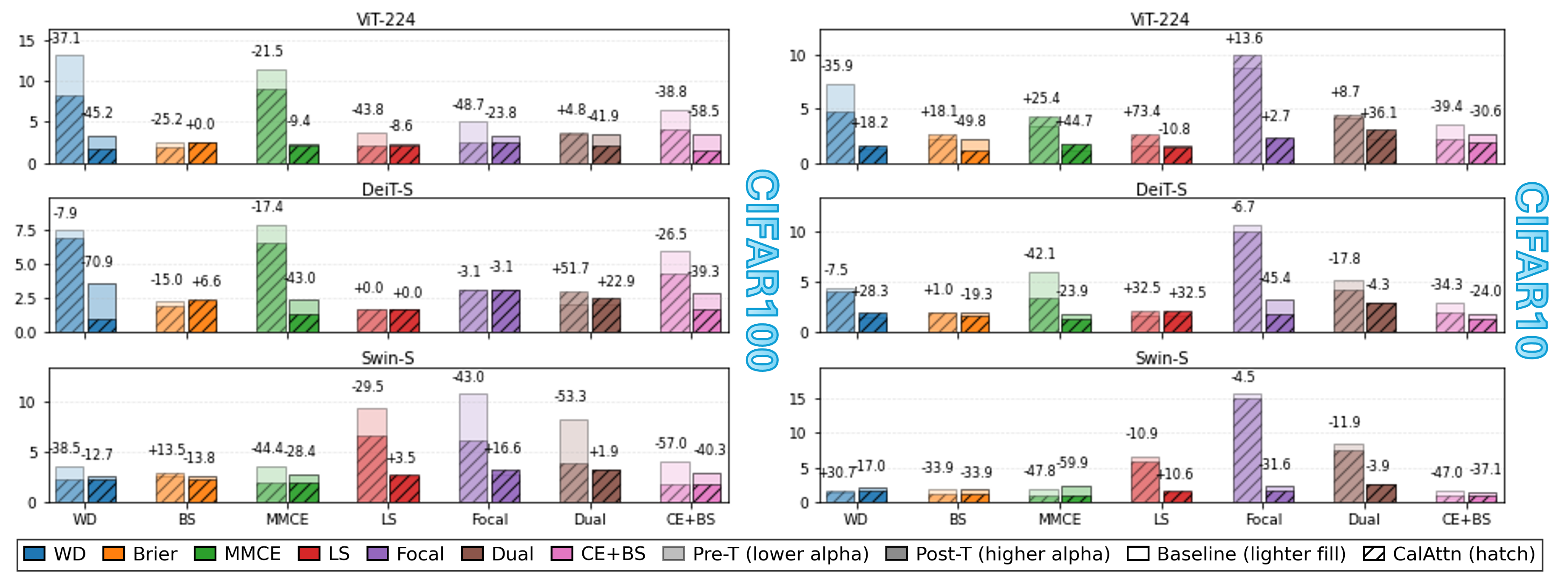}
\caption{AdaECE(\%) on CIFAR-10/100 across ViT-224, DeiT-S, and Swin-S with relative changes ($\Delta$ \%, “$+$” increase, “$-$” decrease).}
\label{fig:delta_AdaECE}
\end{figure*}

We further report Smooth ECE (smECE), Adaptive ECE (AdaECE) and Classwise-ECE in
Tables~\ref{tab:smece_comparison}, ~\ref{tab:adaece_comparison} and~\ref{tab:classece_comparison}. Their reduction of relative changes are reported in Fig.~\ref{fig:delta_smECE} and Fig.~\ref{fig:delta_AdaECE}.
Following standard practice, we use SGD for CNN backbones and AdamW
($\beta=(0.9,0.999)$, weight decay $0.05$) for Transformer backbones
(ViT, DeiT, Swin) with cosine learning-rate scheduling. To ensure that the observed calibration improvements are not artifacts of the
optimizer choice, we additionally retrain each backbone using the alternative
optimizer and evaluate CE, post-hoc TS, and CalAttn under identical settings.
The relative ordering of methods remains unchanged, confirming that the gains
of CalAttn are robust to optimizer variation.

\begin{table*}[h]
\centering
\caption{\(\downarrow\) smECE (\%) before and after applying global temperature scaling.
For each Post-T entry, parentheses report the learned temperature \(T^\star\).
We select \(T^\star\) by minimizing validation ECE with 15 equal-width bins \protect\footnotemark[4].}

\label{tab:smece_comparison}
\resizebox{\textwidth}{!}{%
\begin{tabular}{@{}cccccccccccccccc@{}}
\toprule
\multirow{2}{*}{Dataset} & \multirow{2}{*}{Model} & \multicolumn{2}{c}{\shortstack{Weight Decay \\ \cite{guo2017calibration}}} & \multicolumn{2}{c}{\shortstack{Brier Loss \\ \cite{brier1950verification}}} & \multicolumn{2}{c}{\shortstack{MMCE \\ \cite{kumar2018trainable}}} & \multicolumn{2}{c}{\shortstack{Label Smooth \\ \cite{szegedy2016rethinking}}} & \multicolumn{2}{c}{\shortstack{Focal Loss - 53 \\ \cite{mukhoti2020calibrating}}} & \multicolumn{2}{c}{\shortstack{Dual Focal \\ \cite{tao2023dual}}} & \multicolumn{2}{c}{\shortstack{CE + $\lambda$ BS\\ ($\lambda = 0.1$)} } \\ 
\cmidrule(lr){3-4} \cmidrule(lr){5-6} \cmidrule(lr){7-8} \cmidrule(lr){9-10} \cmidrule(lr){11-12} \cmidrule(lr){13-14} \cmidrule(lr){15-16}
 &  & Pre T & Post-T (T$^\star$) & Pre T & Post-T (T$^\star$) & Pre T & Post-T (T$^\star$) & Pre T & Post-T (T$^\star$) & Pre T & Post-T (T$^\star$) & Pre T & Post-T (T$^\star$) & Pre T & Post-T (T$^\star$) \\ \midrule
\multirow{9}{*}{CIFAR-100} & ResNet-50 & 14.95 & 2.63(2.20) & 5.34 & 3.53(1.10) & 13.40 & 3.12(1.90) & 6.31 & 3.43(1.10)  & 5.56 & 2.37(1.10) & 8.82 & 2.24(1.30) & 13.99 & 2.31(1.60) \\ 
 & ResNet-110 & 15.05 & 3.84(2.30) & 6.53 & 3.51(1.20) & 14.86 & 3.51(2.30)  & 9.55 & 4.36(1.30)  & 10.98 & 3.60(1.30) & 11.69 & 3.31(1.30) & 17.84 & 3.16(1.80) \\ 
 & Wide-ResNet-26-10 & 12.97 & 2.85(2.10) & 4.06 & 2.78(1.10) & 12.27 & 3.85(2.00) & 3.04 & 3.04(1.00) & 2.84 & 2.20(1.10) & 6.01 & 2.41(0.90) & 6.45 & 5.12(1.20) \\ 
 & DenseNet-121 & 15.42 & 2.61(2.20) & 3.66 & 1.83(1.10) & 15.96 & 2.95(2.00) & 3.85 & 3.68(1.10) & 3.04 & \textbf{1.50(1.10)} & 4.07 & 1.72(0.90) & 15.75 & 2.83(1.60) \\ 
& \textbf{ViT\textsubscript{224}} & 13.07 & 3.32(1.30) & 2.62 & 2.25(1.10) & 11.51 & 2.25(1.30) & 3.77 & 2.17(1.10) & 5.13 & 3.20(1.10) & 3.43 & 3.43(1.00) & 6.27 & 2.25(1.50) \\
\rowcolor{yellow!20}
& \textbf{ViT\textsubscript{224}+CalAttn(Ours)} & 8.24 & 1.92(1.20) & \textbf{1.93} & 2.30(1.10) & 9.04 & 2.16(1.20) & 1.97 & 1.97(1.00) & \textbf{2.64} & 2.64(1.00) & 3.66 & 1.96(1.10) & 4.00 & 1.49(1.10) \\
& \textbf{DeiT\textsubscript{small}} & 7.45 & 3.48(1.10) & 2.13 & 2.13(1.00) & 7.88 & 2.34(1.20) & 1.72 & 1.72(1.00) & 3.01 & 3.01(1.00) & \textbf{1.91 }& \textbf{1.91(1.00)} & 5.13 & 2.06(1.40) \\
\rowcolor{yellow!20}
& \textbf{DeiT\textsubscript{small}+CalAttn(Ours)} & 6.87 & \textbf{1.59(1.20)} & 2.13 & 2.05(1.10) & 6.51 & \textbf{1.28(1.20)} & \textbf{1.61} & \textbf{1.61(1.00)} & 3.08 & 3.08(1.00) & 3.05 & 2.16(1.10) & 4.37 & \textbf{1.17(1.10)} \\
& \textbf{Swin\textsubscript{small}} & 3.44 & 2.44(0.90) & 2.55 & 2.55(1.00) & 3.37 & 2.66(0.90) & 8.76 & 2.35(0.80) & 10.55 & 2.42(0.80) & 8.17 & 3.04(0.80) & 3.23 & 2.45(1.20) \\
\rowcolor{yellow!20}
& \textbf{Swin\textsubscript{small}+CalAttn(Ours)} & \textbf{1.99} & 1.99(1.00) & 3.02 & \textbf{1.92(1.10)} &  \textbf{1.76} & 1.76(1.00) & 6.48 & 2.56(0.90) & 5.99 & 2.99(0.90) & 3.76 & 2.92(0.90) & \textbf{1.64} & 1.64(1.00) \\
 
 \midrule
\multirow{9}{*}{CIFAR-10} & ResNet-50 & 3.27 & 1.48(2.50) &1.83 & 1.27(1.10) & 3.41 & 1.47(2.60) & 3.31 & 1.92(0.90) & \textbf{1.45} & 1.45(1.00) & \textbf{1.36} & 1.36(1.00) & 5.18 & 1.67(1.90) \\ 
 & ResNet-110 & 3.43 & 1.88(2.60) & 2.49 & 1.63(1.20) & 3.38 & 1.55(2.80) & 2.87 & 1.52(0.90) & 1.69 & \textbf{1.39(1.10)} & 1.48 & 1.35(1.10) & 5.28 & 2.25(1.90) \\ 
 & Wide-ResNet-26-10 & 2.70 & 1.25(2.20) & 1.55 & 1.55(1.00) & 3.03 & 1.46(2.20) & 3.75 & 1.47(0.80)  & 1.84 & 1.46(0.90) & 3.31 & \textbf{1.20(0.80)} & 3.41 & 1.64(1.50) \\ 
 & DenseNet-121 & 5.56 & 2.21(2.40) & 3.86 & 1.93(1.10) & 5.86 & 1.95(2.30) & 2.84 & 1.63(0.90) & 3.04 & 1.50(0.90) & 4.07 & 1.72(0.90) & 5.17 & 2.24(1.60) \\ 
& \textbf{ViT\textsubscript{224}} & 7.40 & \textbf{1.46(1.30)} & 1.89 & 1.89(1.00) & 3.47 & 1.37(1.10) & 1.66 & 1.66(1.00) & 8.53 & 1.97(0.70) & 3.66 & 2.12(0.90) & 4.26 & 3.53(1.40) \\
\rowcolor{yellow!20}
& \textbf{ViT\textsubscript{224}+CalAttn(Ours)} & 4.75 & 1.54(1.20) & 2.56 & 1.86(1.10) & 4.36 & 1.70(1.20) & 2.41 & \textbf{1.31(0.90)} & 9.95 & 2.06(0.70) & 4.35 & 2.78(0.90) & 1.57 & 1.18(1.10) \\
& \textbf{DeiT\textsubscript{small}} & 4.41 & 1.56(1.20) & 2.03 & 2.03(1.00) & 6.01 & 1.65(1.20) & \textbf{1.64} & 1.64(1.00) & 10.49 & 2.87(0.70) & 5.00 & 2.79(0.80) & 3.85 & 1.96(1.30) \\
\rowcolor{yellow!20}
& \textbf{DeiT\textsubscript{small}+CalAttn(Ours)} & 4.07 & 1.76(1.10) & 2.09 & 1.76(1.10) & 3.39 & 1.41(1.10) & 2.15 & 2.15(1.00) & 9.90 & 1.79(0.70) & 3.77 & 2.74(0.90) & 1.89 & 1.36(1.10) \\
& \textbf{Swin\textsubscript{small}} & \textbf{1.43} & 1.93(0.90) & 1.46 & 1.46(1.00) & 1.78 & 2.31(0.90) & 6.68 & 1.36(0.80) & 15.63 & 2.34(0.60) & 8.40 & 2.41(0.70) & 3.11 & 2.04(1.20) \\
\rowcolor{yellow!20}
& \textbf{Swin\textsubscript{small}+CalAttn(Ours)} & 1.63 & 1.63(1.00) & \textbf{1.19} & \textbf{1.19(1.00)} & \textbf{1.23} & \textbf{1.23(1.00)} & 5.94 & 1.55(0.80) & 14.94 & 1.61(0.60) & 7.08 & 2.32(0.80) &\textbf{1.11} & \textbf{1.11(1.00)}  \\

\midrule
\multirow{5}{*}{MNIST} & \textbf{ViT\textsubscript{224}} & 2.88 & 1.13(0.90) & 39.17 & 3.72(8.70) & 19.77 & 4.89(2.10) & 12.74 & 6.61(1.50) & 7.13 & 2.15(1.40) & 18.28 & 4.52(2.30) & 5.23 & 1.26(0.80) \\
\rowcolor{yellow!20}
& \textbf{ViT\textsubscript{224}+CalAttn(Ours)} & 2.76 & 1.13(0.90) & 3.12 & 1.04(0.90) & 3.89 & 1.41(0.80) & 8.30 & 1.31(0.70) & 23.30 & 1.25(0.40) & 17.99 & 1.45(0.50) & 0.94 & 0.68(0.90) \\
& \textbf{DeiT\textsubscript{small}} & 2.87 & 1.29(0.90) & 26.66 & 2.53(3.20) & 4.56 & 1.17(0.80) & 8.70 & 1.05(0.70) & 21.32 & 2.39(0.50) &15.89 & 1.65(0.70) & 4.45 & 1.30(0.80) \\
\rowcolor{yellow!20}
& \textbf{DeiT\textsubscript{small}+CalAttn(Ours)} & 1.23 & 1.23(1.00) & 1.17 & 0.68(0.90) & 2.15 & 1.13(0.90) & \textbf{6.21} & 1.40(0.80) & 18.17 & 1.56(0.50) &  8.96 & 2.30(0.60) & 1.14 & 1.14(1.00) \\
& \textbf{Swin\textsubscript{small}} & 3.57 & 3.20(1.10) & 1.28 & 0.71(0.90) & 1.35 & 0.79(0.90) & 8.12 & 0.79(0.70) & 11.77 & 0.96(0.50) & \textbf{5.06} & 0.68(0.60) & 0.65 & 0.65(1.00) \\
\rowcolor{yellow!20}
& \textbf{Swin\textsubscript{small}+CalAttn(Ours)} & \textbf{0.53} & \textbf{0.66(1.10)} & \textbf{0.90} & \textbf{0.59(0.90)} & \textbf{1.22} & \textbf{0.68(0.90)} & 7.89 & \textbf{0.67(0.70)} & \textbf{7.11} & \textbf{0.58(0.50)} & 8.81 & \textbf{0.66(0.50)} & \textbf{0.90} & \textbf{0.59(0.90)} \\

\midrule
\multirow{5}{*}{Tiny-ImageNet} 
& \textbf{ViT\textsubscript{224}} & 4.39 & 1.80(1.10) & 4.40 & 1.81(1.10) & 4.96 & 1.83(1.20) & 2.20 & 1.65(1.10) & 3.19 & 1.85(1.10) & 2.35 & 2.33(1.10) & 2.45 & 2.41(1.10) \\
\rowcolor{yellow!20}
& \textbf{ViT\textsubscript{224}+CalAttn(Ours)} & 21.61 & 2.88(1.90) & 9.17 & 0.97(1.30) & 19.59 & 2.95(1.80) & 2.00 & 1.59(1.10) & \textbf{1.68} & \textbf{0.14(1.30)} & 13.52 & 1.44(1.40) & \textbf{0.89} & 0.89(1.00) \\

& \textbf{DeiT\textsubscript{small}} &13.00  & 1.96(1.90) & 2.53 & 1.81(1.10) & 13.61 & \textbf{1.16(1.50)} & \textbf{1.65} & 1.65(1.00) & 10.25 & 2.18(1.70) & 9.36 & 1.90(1.70) & 1.88 & 1.86(1.10) \\
\rowcolor{yellow!20}
& \textbf{DeiT\textsubscript{small}+CalAttn(Ours)} & \textbf{2.29} & 1.44(1.40) & 2.32 & 1.12(1.20) & 2.72 & 1.45(1.50) & 2.28 & \textbf{1.22(1.10)} & 2.18 & 1.36(1.30) & \textbf{1.90} & \textbf{1.19(1.30)} & 1.18 & \textbf{0.67(1.20)} \\
& \textbf{Swin\textsubscript{small}} & 3.24 & 2.53(0.90) & 0.88 & 0.66(0.90) & 2.04 & 2.04(1.00) & 4.79 & 2.34(0.90) & 4.20 & 2.45(0.90) & 4.64 & 2.46(0.90) & 1.60 & 1.60(1.00) \\
\rowcolor{yellow!20}
& \textbf{Swin\textsubscript{small}+CalAttn(Ours)} & 2.88 & \textbf{1.32(1.10)} & \textbf{0.86} & \textbf{0.78(1.10)} & \textbf{1.50} & 1.50(1.00) & 2.51 & 2.51(1.00) & 2.31 & 2.31(1.00) & 2.35 & 2.35(1.00) & 0.92 & 0.64(0.90)\\

\midrule
\bottomrule
\end{tabular}
}
\end{table*}

\footnotetext[4]{Smooth ECE (smECE) is introduced by~\cite{blasiok2023smooth}.
We use the authors' public implementation~\cite{smce-github-url} with bandwidth $h=0.05$.}

\section{Impact of Calibration on Model Size}
\label{app:model_size}

\paragraph{Computational and parameter overhead.}
CalAttn adds a two-layer MLP that maps the final \texttt{[CLS]} embedding of
dimension $d$ to a scalar scale using hidden width $h$. The additional parameter
count is
\[
\underbrace{hd}_{W_1}+\underbrace{h}_{b_1}+\underbrace{h}_{w_2}+\underbrace{1}_{b_2}
= O(dh),
\]
and the extra per-sample compute is dominated by two matrix-vector products,
i.e., $O(dh)$ FLOPs, plus a scalar division. This overhead is negligible
relative to a Transformer encoder, whose attention and MLP blocks scale on the
order of $O(LN d^2)$ with depth $L$ and token count $N$.

\begin{table}[t]
\centering
\caption{\textbf{Checkpoint size overhead of CalAttn.}
We report the on-disk checkpoint size (MB) of baseline models and models
augmented with CalAttn under the same serialization settings; $\Delta$ denotes
the absolute increase.}
\label{tab:model_size}
\small
\setlength{\tabcolsep}{5pt}
\renewcommand{\arraystretch}{1.1}
\begin{tabular}{lcccc}
\toprule
\textbf{Dataset} & \textbf{Backbone} & \textbf{Variant} & \textbf{Size (MB)} & $\Delta$ (MB) \\
\midrule
\multirow{6}{*}{C-10}
 & ViT\textsubscript{224} & baseline         & 327.38 & -- \\
 &                        & + CalAttn (ours) & 327.76 & +0.38 \\
 & DeiT\textsubscript{small} & baseline      &  82.72 & -- \\
 &                        & + CalAttn (ours) &  82.91 & +0.19 \\
 & Swin\textsubscript{small} & baseline      & 186.16 & -- \\
 &                        & + CalAttn (ours) & 186.54 & +0.38 \\
\addlinespace
\multirow{6}{*}{C-100}
 & ViT\textsubscript{224} & baseline         & 327.64 & -- \\
 &                        & + CalAttn (ours) & 328.02 & +0.38 \\
 & DeiT\textsubscript{small} & baseline      &  82.85 & -- \\
 &                        & + CalAttn (ours) &  83.04 & +0.19 \\
 & Swin\textsubscript{small} & baseline      & 186.42 & -- \\
 &                        & + CalAttn (ours) & 186.80 & +0.38 \\
\bottomrule
\end{tabular}
\end{table}


\section{Distinguishing CalAttn from SATS}
\label{app:calattn_vs_sats}

Although both SATS~\cite{joy2023sample} and CalAttn predict an instance-wise
temperature, they optimize different objectives and intervene at different
stages of the model.

\paragraph{Optimization: post-hoc regression vs.\ end-to-end learning.}
SATS is a \emph{post-hoc} method: the backbone is frozen and a temperature
regressor is fit on a held-out validation set to minimize a calibration
criterion (e.g., NLL or ECE):
\[
\min_{\phi}\; \mathbb{E}_{(x,y)\sim\mathcal D_{\mathrm{val}}}\,
\mathcal{C}\!\left(
\mathrm{softmax}\!\left(\tfrac{\boldsymbol{\ell}(x)}{T_\phi(\boldsymbol{\ell}(x))}\right),\, y
\right),
\]
where logits $\boldsymbol{\ell}(x)$ are produced by a fixed model.

CalAttn is integrated into training and jointly optimizes the backbone and the
scale head under a proper scoring objective:
\[
\min_{\theta,\psi}\;
\mathbb{E}_{(x,y)\sim\mathcal D}\!\left[
\mathrm{CE}\!\left(\mathrm{softmax}\!\left(\tfrac{\boldsymbol{\ell}_\theta(x)}
{s_\psi(z_\theta(x))}\right),\,y\right)
+\lambda\,\bigl\|\widehat{\mathbf y}-\mathbf e_y\bigr\|_2^2
\right],
\]
where $z_\theta(x)$ denotes the final \texttt{[CLS]} embedding and
$\widehat{\mathbf y}=\mathrm{softmax}(\boldsymbol{\ell}_\theta(x)/s_\psi(z_\theta(x)))$.

\paragraph{Conditioning signal: logits vs.\ representation.}
SATS conditions temperature on the \emph{logits} via $T_\phi(\boldsymbol{\ell}(x))$.
Since logits are the output of the classifier head, SATS performs confidence
correction \emph{after} the representation and decision boundary have been
learned, and it cannot influence feature learning.

In contrast, CalAttn predicts the scale from an \emph{upstream} representation,
$s_\psi(z_\theta(x))$, before the classifier head. Consequently, gradients from
the calibration objective backpropagate through $s_\psi(\cdot)$ into the
\texttt{[CLS]} pathway, enabling representation learning to co-adapt with
instance-wise confidence modulation. This distinction provides a different
inductive bias from post-hoc regression on frozen outputs and helps explain why
CalAttn can remain complementary to global post-hoc temperature scaling.

\section{Sensitivity to the CE+Brier Trade-off $\lambda$ (CIFAR-100, DeiT-S)}
\label{app:lambda_sensitivity}

Unless stated otherwise, all metrics are reported in \% and computed \emph{after}
applying a single global post-hoc temperature scaling step (TS) to each trained model.

\begin{table}[t]
\centering
\caption{DeiT-S trained with CE+$\lambda$Brier (without CalAttn), evaluated after TS.}
\label{tab:lambda_cebs}
\small
\setlength{\tabcolsep}{4pt}
\renewcommand{\arraystretch}{1.1}
\begin{tabular}{ccccc}
\toprule
$\lambda$ & ECE $\downarrow$ & AdaECE $\downarrow$ & ClassECE $\downarrow$ & smECE $\downarrow$ \\
\midrule
0.1 & \textbf{2.39} & 2.44 & \textbf{0.26} & \textbf{2.44} \\
0.2 & 2.48 & \textbf{2.38} & 0.27 & 2.47 \\
0.3 & 2.50 & 2.39 & 0.27 & 2.48 \\
0.4 & 2.48 & \textbf{2.38} & 0.28 & 2.47 \\
0.5--1.0 & 2.48 & \textbf{2.38} & 0.27 & 2.47 \\
\bottomrule
\end{tabular}
\end{table}

\begin{table}[t]
\centering
\caption{DeiT-S trained with CE+$\lambda$Brier+CalAttn, evaluated after TS.}
\label{tab:lambda_full}
\small
\setlength{\tabcolsep}{4pt}
\renewcommand{\arraystretch}{1.1}
\begin{tabular}{ccccc}
\toprule
$\lambda$ & ECE $\downarrow$ & AdaECE $\downarrow$ & ClassECE $\downarrow$ & smECE $\downarrow$ \\
\midrule
0.1 & 2.03 & 1.97 & 0.26 & 1.98 \\
0.3--0.8 & 2.30 & 2.30 & 0.25 & 2.20 \\
0.9 & \textbf{1.43} & \textbf{1.51} & 0.25 & \textbf{1.44} \\
1.0 & 2.36 & 2.34 & 0.25 & 2.24 \\
\bottomrule
\end{tabular}
\end{table}

\paragraph{Takeaways.}
\textbf{(1) CE+$\lambda$Brier without CalAttn is weakly sensitive to $\lambda$.}
ECE varies by at most $0.11$ percentage points over the sweep, indicating a
broad plateau.

\textbf{(2) With CalAttn, performance depends more on $\lambda$ but remains stable near the default.}
The best calibration in this sweep occurs at $\lambda=0.9$; however,
$\lambda=0.1$ remains within $0.6$ percentage points in ECE, making it a
reasonable default that avoids dataset-specific tuning.




\section{Additional Ablations: ViT-based SATS}
\label{app:sats-vit}

\subsection{Experimental Protocol}

\paragraph{Backbones and datasets.}
We use the same ViT backbones and training recipes as in the main paper and
report results on \textbf{CIFAR-10} and \textbf{CIFAR-100}.
Unless otherwise stated, all models are trained for 350 epochs with identical
data augmentation, optimisers, and learning-rate schedules. \paragraph{Additional baselines.}
Our focus is a ranking-preserving, lightweight scalar calibrator. More expressive post-hoc alternatives (e.g., classwise or vector scaling) can correct class-conditional biases at the expense of additional degrees of freedom and potential decision changes. We include an extended-head comparison (e.g., Dirichlet-style parameterization) to illustrate this capacity--complexity trade-off and leave broader calibrator families and ensembles as future work.

\paragraph{Calibration variants.}
We compare the following methods:
(i) \textbf{ViT (baseline)} trained with cross-entropy;
(ii) \textbf{ViT+CalAttn} (our representation-conditioned temperature head with Brier penalty);
(iii) \textbf{ViT+CalAttn+TS} (post-hoc global temperature scaling on CalAttn);
(iv) \textbf{ViT+CalAttn+SATS}, where SATS is applied post-hoc on top of CalAttn.
SATS follows \cite{joy2023sample}, using a lightweight MLP that predicts a
per-sample temperature from logits (and optionally the CLS embedding).

\paragraph{Temperature scaling.}
Global TS selects $T\in\{0.1,0.2,\dots,10.0\}$ on a held-out $5\%$ validation split
to minimise \textbf{ECE}, following \cite{mukhoti2020calibrating} for
comparability with focal-family baselines.

\paragraph{SATS training.}
SATS is trained post-hoc on the same validation split for 10 epochs with early
stopping on validation ECE. Unless stated otherwise, the input consists only of
pre-softmax logits, matching \cite{joy2023sample}.

\paragraph{Metrics.}
We report mean$\pm$std over three random seeds (0,1,2) for Top-1 accuracy,
negative log-likelihood (NLL), Brier score, ECE, and smECE (bandwidth $h=0.05$).

\subsection{Results}
\label{app:sats-results}

Across both datasets, CalAttn consistently reduces ECE and smECE relative
to the ViT baseline without affecting Top-1 accuracy. Adding a global TS step on
top of CalAttn yields the strongest improvements in all calibration metrics.
SATS-only improves calibration over the baseline, but underperforms
CalAttn+TS, indicating that learning an representation-conditioned temperature \emph{during}
training provides a stronger inductive bias than purely post-hoc fitting. This suggests that calibrating representations during training induces a smoother uncertainty geometry than fitting a temperature regressor post-hoc, which may overfit the validation split.
\begin{table*}[h]
\centering
\caption{\textbf{ViT (3 seeds).} Mean$\pm$std on CIFAR-100 and CIFAR-10. Lower is better for losses and calibration.}
\label{tab:c100&10_vit_3seeds}
\setlength{\tabcolsep}{4pt}
\resizebox{\textwidth}{!}{%
\begin{tabular}{lccccc|ccccc}
\toprule
 & \multicolumn{5}{c}{\textbf{CIFAR-100}} & \multicolumn{5}{c}{\textbf{CIFAR-10}} \\
\cmidrule(lr){2-6} \cmidrule(lr){7-11}
Method 
& Top-1 & NLL & Brier & ECE & smECE
& Top-1 & NLL & Brier & ECE & smECE \\
\midrule
Baseline
& $66.01\!\pm\!0.22$ & $2.608\!\pm\!0.012$ & $0.791\!\pm\!0.004$ & $8.54\!\pm\!0.26$ & $8.63\!\pm\!0.24$
& $77.20\!\pm\!0.31$ & $0.850\!\pm\!0.020$ & $0.224\!\pm\!0.004$ & $3.35\!\pm\!0.20$ & $3.52\!\pm\!0.22$ \\

CalAttn
& $66.20\!\pm\!0.35$ & $2.514\!\pm\!0.014$ & $0.779\!\pm\!0.004$ & $6.39\!\pm\!0.21$ & $6.48\!\pm\!0.23$
& $77.50\!\pm\!0.20$ & $0.842\!\pm\!0.018$ & $0.210\!\pm\!0.003$ & $2.10\!\pm\!0.15$ & $2.25\!\pm\!0.16$ \\

CalAttn+TS
& $67.20\!\pm\!0.26$ & $\mathbf{2.489\!\pm\!0.011}$ & $\mathbf{0.774\!\pm\!0.003}$ & $\mathbf{1.42\!\pm\!0.12}$ & $\mathbf{1.10\!\pm\!0.14}$
& $78.10\!\pm\!0.25$ & $\mathbf{0.792\!\pm\!0.017}$ & $\mathbf{0.202\!\pm\!0.003}$ & $\mathbf{0.92\!\pm\!0.08}$ & $\mathbf{1.01\!\pm\!0.09}$ \\

CalAttn+SATS
& $66.80\!\pm\!0.28$ & $2.519\!\pm\!0.015$ & $0.780\!\pm\!0.004$ & $2.70\!\pm\!0.15$ & $3.44\!\pm\!0.20$
& $77.70\!\pm\!0.34$ & $0.812\!\pm\!0.019$ & $0.206\!\pm\!0.003$ & $1.12\!\pm\!0.10$ & $1.20\!\pm\!0.11$ \\
\bottomrule
\end{tabular}}
\end{table*}

\begin{table*}[h]
\centering
\caption{\(\downarrow\) AdaECE (\%) before and after applying global temperature scaling.
We select \(T^\star\) by minimizing validation ECE with 15 equal-width bins; parentheses report \(T^\star\).
AdaECE is computed with 15 adaptive (equal-count) bins.}

\label{tab:adaece_comparison}
\resizebox{\textwidth}{!}{%
\begin{tabular}{@{}cccccccccccccccc@{}}
\toprule
\multirow{2}{*}{Dataset} & \multirow{2}{*}{Model} & \multicolumn{2}{c}{\shortstack{Weight Decay \\ \cite{guo2017calibration}}} & \multicolumn{2}{c}{\shortstack{Brier Loss \\ \cite{brier1950verification}}} & \multicolumn{2}{c}{\shortstack{MMCE \\ \cite{kumar2018trainable}}} & \multicolumn{2}{c}{\shortstack{Label Smooth \\ \cite{szegedy2016rethinking}}} & \multicolumn{2}{c}{\shortstack{Focal Loss - 53 \\ \cite{mukhoti2020calibrating}}} & \multicolumn{2}{c}{\shortstack{Dual Focal \\ \cite{tao2023dual}}} & \multicolumn{2}{c}{\shortstack{CE + $\lambda$ BS\\ ($\lambda = 0.1$)} } \\ 
\cmidrule(lr){3-4} \cmidrule(lr){5-6} \cmidrule(lr){7-8} \cmidrule(lr){9-10} \cmidrule(lr){11-12} \cmidrule(lr){13-14} \cmidrule(lr){15-16}
 &  & Pre T & Post-T (T$^\star$) & Pre T & Post-T (T$^\star$) & Pre T & Post-T (T$^\star$) & Pre T & Post-T (T$^\star$) & Pre T &Post-T (T$^\star$) & Pre T &Post-T (T$^\star$) & Pre T &Post-T (T$^\star$) \\ \midrule
\multirow{10}{*}{CIFAR-100} & ResNet-50 & 17.99 & 3.38(2.20) & 5.46 & 4.24(1.10) & 15.05 & 3.42(1.90) & 6.72 & 5.37(1.10)  & 5.64 & 2.84(1.10) & 8.79 & 2.27(1.30) & 14.41 & 2.38(1.60) \\ 
 & ResNet-110 & 19.28 & 6.27(2.30) & 6.51 & 3.75(1.20) & 18.83 & 4.86(2.30)  & 9.68 & 8.11(1.30)  & 10.90 & 4.13(1.30) & 11.64 & 4.48(1.30) & 19.35 & 3.51(1.80) \\ 
 & Wide-ResNet-26-10 & 15.16 & 3.23(2.10) & 4.08 & 3.11(1.10) & 13.55 & 3.83(2.00) & 3.73 & 3.73(1.00) & \textbf{2.40} & 2.38(1.10) & 5.36 & 2.38(1.20) & 6.59 & 5.19(1.20) \\ 
 & DenseNet-121 & 19.07 & 3.82(2.20) & 3.92 & 2.41(1.10) & 17.37 & 3.07(2.00) & 8.62 & 5.92(1.10) & 3.35 & 1.80(1.10) & 6.69 & \textbf{1.69(1.20)} & 16.09 & 3.04(1.60) \\ 
& \textbf{ViT\textsubscript{224}} & 13.11 & 3.34(1.30) & \textbf{2.58} & 2.56(1.10) & 11.51 & 2.33(1.30) & 3.77 & \textbf{2.32(1.10)} & 5.07 & 3.41(1.10) & 3.56 & 3.56(1.00) & 6.55 & 3.54(1.50) \\
\rowcolor{yellow!20}
& \textbf{ViT\textsubscript{224}+CalAttn(Ours)} & 8.25 & 1.83(1.20) & \textbf{1.93} & 2.56(1.10) & 9.04 & 2.11(1.20) & 2.12 & 2.12(1.00) & 2.60 & 2.60(1.00) & 3.73 & 2.07(1.10) & 4.01 & \textbf{1.47(1.10)} \\
& \textbf{DeiT\textsubscript{small}} & 7.46 & 3.57(1.10) & 2.26 & 2.26(1.00) & 7.88 & 2.35(1.20) & \textbf{1.67} & \textbf{1.67(1.00)} & 3.18 & 3.18(1.00) & 2.01 & 2.01(1.00) & 5.93 & 2.85(1.40) \\
\rowcolor{yellow!20}
& \textbf{DeiT\textsubscript{small}+CalAttn(Ours)} & 6.87 & \textbf{1.04(1.20)} & 1.92 & 2.41(1.10) & 6.51 & 1.34(1.20) & \textbf{1.67} & 1.67(1.00) & 3.08 & 3.08(1.00) & 3.05 & 2.47(1.10) & 4.36 & 1.73(1.10) \\
& \textbf{Swin\textsubscript{small}} & 3.58 & 2.52(0.90) & \textbf{2.60} & 2.60(1.00) & 3.54 & 2.75(0.90) & 9.29 & \textbf{2.56(0.80)} & 10.71 & 2.71(0.80) & 8.24 & 3.16(0.80) & 4.07 & 2.93(1.20) \\
\rowcolor{yellow!20}
& \textbf{Swin\textsubscript{small}+CalAttn(Ours)} & 2.20 & 2.20(1.00) & 2.95 & 2.24(1.10) & 1.97 & 1.97(1.00) & 6.55 & 2.65(0.90) & 6.10 & 3.16(0.90) & 3.85 & 3.22(0.90) & \textbf{1.75} & \textbf{1.75(1.00)} \\

 \midrule
\multirow{8}{*}{CIFAR-10} & ResNet-50 & 4.22 & 2.11(2.50) & 1.85 & 1.34(1.10) & 4.67 & 2.01(2.60) & 4.28 & 3.20(0.90) & 1.64 & 1.64(1.00) & 1.28 & 1.28(1.00) & 5.69 & 1.98(1.90) \\ 
 & ResNet-110 & 4.78 & 2.42(2.60) & 2.52 & 1.72(1.20) & 5.21 & 2.66(2.80) & 4.57 & 3.62(0.90) & 1.76 & 1.32(1.10) & 1.69 & 1.42(1.10) & 6.23 & 2.88(1.90) \\ 
 & Wide-ResNet-26-10 & 3.22 & 1.62(2.20) & 1.94 & 1.94(1.00) & 3.58 & 1.83(2.20) & 4.58 & 2.55(0.80)  & 1.84 & 1.63(0.90) & 3.16 & \textbf{1.20(0.80)} & 3.41 & 1.76(1.50) \\ 
 & DenseNet-121 & 4.69 & 2.28(2.40) & 1.84 & 1.84(1.00) & 4.97 & 2.69(2.40) & 4.60 & 3.36(0.90) & 1.58 & 1.62(0.90) & \textbf{0.79} & 1.32(0.90) & 5.12 & 3.25(1.60) \\ 
& \textbf{ViT\textsubscript{224}} & 7.39 & 1.37(1.30) & 2.27 & 2.27(1.00) & 3.47 & 1.23(1.10) & \textbf{1.58} & \textbf{1.58(1.00)} & 8.77 & 2.24(0.70) & 4.16 & 2.27(0.90) & 3.58 & 2.68(1.40) \\
\rowcolor{yellow!20}
& \textbf{ViT\textsubscript{224}+CalAttn(Ours)} & 4.74 & 1.62(1.20) & 2.68 & \textbf{1.14(1.10)} & 4.35 & 1.78(1.20) & 2.74 & 1.41(0.90) & 9.96 & 2.30(0.70) & 4.52 & 3.09(0.90) & \textbf{2.17} & 1.86(1.10) \\
& \textbf{DeiT\textsubscript{small}} & 4.41 & \textbf{1.52(1.20)} & 1.97 & 1.97(1.00) & 6.01 & 1.84(1.20) & \textbf{1.66} & 1.66(1.00) & 10.62 & 3.17(0.70) & 5.11 & 3.00(0.80) & 2.97 & 1.83(1.30) \\
\rowcolor{yellow!20}
& \textbf{DeiT\textsubscript{small}+CalAttn(Ours)} & 4.08 & 1.95(1.10) & 1.99 & 1.59(1.10) & 3.48 & 1.40(1.10) & 2.20 & 2.20(1.00) & 9.91 & 1.73(0.70) & 4.20 & 2.87(0.90) & \textbf{1.95} & \textbf{1.39(1.10)} \\
& \textbf{Swin\textsubscript{small}} & \textbf{1.27} & 2.00(0.90) & 1.77 & 1.77(1.00) & 1.86 & 2.42(0.90) & 6.67 & \textbf{1.41(0.80)} & 15.69 & 2.28(0.60) & 8.40 & 2.57(0.70) & 1.66 & 1.40(1.20) \\
\rowcolor{yellow!20}
& \textbf{Swin\textsubscript{small}+CalAttn(Ours)} & 1.66 & 1.66(1.00) & 1.17 & 1.17(1.00) & 0.97 & 0.97(1.00) & 5.94 & 1.56(0.80) & 14.98 & 1.56(0.60) & 7.40 & 2.47(0.80) &\textbf{0.88} & \textbf{0.88(1.00)} \\

\midrule
\multirow{5}{*}{MNIST} & \textbf{ViT\textsubscript{224}} &\textbf{ 2.97} & 1.15(0.90) & 48.83 & 6.07(8.70) & 19.96 & 5.28(2.10) & 12.82 & 8.53(1.50) & 24.06 & 2.41(1.40) & 18.12 & 5.07(2.30) & 5.34 & \textbf{1.08(0.80)} \\
\rowcolor{yellow!20}
& \textbf{ViT\textsubscript{224}+CalAttn(Ours)} & 2.65 & 0.80(0.90) & 3.12 & 0.77(0.90) & 3.89 & 1.08(0.80) & 8.29 & 1.15(0.70) & 7.04 & 1.10(0.40) & 18.38 & 1.54(0.50) & \textbf{0.74} & \textbf{0.44(0.90)} \\
& \textbf{DeiT\textsubscript{small}} & \textbf{2.85} & 1.11(0.90) & 28.36 & 3.06(3.20) & 4.53 & 0.98(0.80) & 8.69 & 0.97(0.70) & 21.72 & 2.26(0.50) & 8.96 & 1.87(0.70) & 4.48 & \textbf{1.04(0.80)} \\
\rowcolor{yellow!20}
& \textbf{DeiT\textsubscript{small}+CalAttn(Ours)} & 1.17 & 1.17(1.00) & 1.08 & 1.04(0.90) & 2.22 & 1.02(0.90) & 6.20 & 1.33(0.80) & 18.28 & 1.81(0.50) & 15.90 & 2.38(0.60) & \textbf{0.91} & \textbf{0.91(1.00)} \\
& \textbf{Swin\textsubscript{small}} & 3.64 & 3.21(1.10) & 1.12 & 0.38(0.90) & 1.24 & 0.50(0.90) & 7.99 & 1.21(0.70) & 7.08 & 0.35(0.50) & 5.04 & 0.37(0.60) & \textbf{0.46} & \textbf{0.46(1.00)} \\
\rowcolor{yellow!20}
& \textbf{Swin\textsubscript{small}+CalAttn(Ours)} & 0.84 & 0.39(1.20)  & 0.76 & 0.32(0.90) & 1.19 & 0.46(0.90) & 6.59 & 1.01(0.70) & 11.74 & 0.75(0.50) & 8.79 & 0.40(0.50) & 0.85 & \textbf{0.42(0.90)} \\

\midrule
\multirow{5}{*}{Tiny-ImageNet} 
& \textbf{ViT\textsubscript{224}} & 4.36 & 1.73(1.10) & 4.40 & \textbf{1.67(1.10)} & 4.96 & 2.39(1.20) & \textbf{2.09} & 2.28(1.10) & 3.23 & 2.23(1.10) & 2.50 & 2.68(1.10) & 1.44 & 1.39(1.10) \\
\rowcolor{yellow!20}
& \textbf{ViT\textsubscript{224}+CalAttn(Ours)} & 22.11 & 3.01(1.90) & 9.18 & 0.85(1.30) & 19.84 & 3.02(1.80) & 2.05 & 1.98(1.10) & 1.68 & 1.23(1.10) & 13.57 & 1.10(1.40) & \textbf{1.01} & \textbf{0.78(1.10)} \\

& \textbf{DeiT\textsubscript{small}} & 13.04 & 2.25(1.40) & 2.38 & 2.04(1.10) & 13.66 & 1.37(1.50) & \textbf{1.71} & \textbf{1.71(1.00)} & 10.26 & 2.39(1.70) & 9.37 & 1.86(1.50) &  1.43 & 1.29(1.10) \\
\rowcolor{yellow!20}
& \textbf{DeiT\textsubscript{small}+CalAttn(Ours)} & 2.12 & 1.77(1.40) & 2.26 & 1.47(1.20)  & 2.61 & 1.07(1.50) & 2.29 & 1.35(1.10) & 2.39 & 1.26(1.30) &  1.86 & 1.41(1.30) & \textbf{0.95} & \textbf{0.83(1.20)} \\
& \textbf{Swin\textsubscript{small}} & 3.48 & 2.54(0.90) & \textbf{1.05} & \textbf{0.54(0.90)} & 2.15 & 2.15(1.00) & 4.98 & 2.44(0.90) & 4.66 & 2.69(0.90) & 4.96 & 2.62(0.90) &  0.91 & 0.91(1.00) \\
\rowcolor{yellow!20}
& \textbf{Swin\textsubscript{small}+CalAttn(Ours)} & 2.84 & 1.63(1.10) &\textbf{0.84} & 1.12(1.10) & 1.57 & 1.57(1.00) & 2.62 & 2.62(1.00) & 2.37 & 2.37(1.00) & 2.29 & 2.29(1.00) & 0.82 & \textbf{0.24(0.90)} \\

\midrule
\bottomrule
\end{tabular}
}
\end{table*}

\begin{table*}[h]
\centering
\caption{\(\downarrow\) Classwise-ECE (\%) before and after applying global temperature scaling (15 bins).}
\label{tab:classece_comparison}
\resizebox{\textwidth}{!}{%
\begin{tabular}{@{}cccccccccccccccc@{}}
\toprule
\multirow{2}{*}{Dataset} & \multirow{2}{*}{Model} & \multicolumn{2}{c}{\shortstack{Weight Decay \\ \cite{guo2017calibration}}} & \multicolumn{2}{c}{\shortstack{Brier Loss \\ \cite{brier1950verification}}} & \multicolumn{2}{c}{\shortstack{MMCE \\ \cite{kumar2018trainable}}} & \multicolumn{2}{c}{\shortstack{Label Smooth \\ \cite{szegedy2016rethinking}}} & \multicolumn{2}{c}{\shortstack{Focal Loss - 53 \\ \cite{mukhoti2020calibrating}}} & \multicolumn{2}{c}{\shortstack{Dual Focal \\ \cite{tao2023dual}}} & \multicolumn{2}{c}{\shortstack{CE + $\lambda$ BS\\ ($\lambda = 0.1$)} } \\ 
\cmidrule(lr){3-4} \cmidrule(lr){5-6} \cmidrule(lr){7-8} \cmidrule(lr){9-10} \cmidrule(lr){11-12} \cmidrule(lr){13-14} \cmidrule(lr){15-16}
 &  & Pre T & Post-T (T$^\star$) & Pre T & Post-T (T$^\star$) & Pre T & Post-T (T$^\star$) & Pre T & Post-T (T$^\star$) & Pre T & Post-T (T$^\star$) & Pre T & Post-T (T$^\star$) & Pre T & Post-T (T$^\star$) \\ \midrule
\multirow{10}{*}{CIFAR-100} & ResNet-50 & 0.39 & 0.22(2.20) & 0.21 & 0.22(1.10) & 0.34 & 0.22(1.90) & 0.21 & 0.22(1.10)  & 0.21 & 0.20(1.10) & 0.24 & 0.22(1.30) & 0.34 & 0.23(1.60)\\ 
 & ResNet-110 & 0.42 & 0.21(2.30) & 0.22 & 0.23(1.20) & 0.41 & 0.22(2.30)  & 0.25 & 0.23(1.30)  & 0.27 & 0.22(1.30) & 0.28 & 0.21(1.30) & 0.43 & 0.23(1.80)\\ 
 & Wide-ResNet-26-10 & 0.34 & 0.21(2.10) & 0.19 & 0.20(1.10) & 0.31 & 0.21(2.00) & 0.20 & 0.20(1.00) &\textbf{0.18} & \textbf{0.20(1.10)} & 0.19 & 0.20(1.20) & 0.33 & 0.30(1.20)\\ 
 & DenseNet-121 & 0.42 & 0.22(2.20) & 0.21 & 0.21(1.10) & 0.39 & 0.23(2.00) & 0.23 & 0.21(1.10) & 0.20 & 0.21(1.10) & 0.22 & 0.21(1.20) & 0.37 & 0.25(1.60)\\ 
& \textbf{ViT\textsubscript{224}} & 0.43 & 0.26(1.30) & 0.29 & 0.28(1.10) & 0.40 & 0.26(1.30) & \textbf{0.27} & \textbf{0.25(1.10)} & 0.33 & 0.29(1.10) & 0.28 & 0.28(1.00) & 0.30 & 0.27(1.50) \\
\rowcolor{yellow!20}
& \textbf{ViT\textsubscript{224}+CalAttn(Ours)} & 0.34 & 0.27(1.20) & 0.35 & 0.34(1.10) & 0.36 & 0.29(1.20) & \textbf{0.28 }& 0.28(1.00) & 0.30 & 0.30(1.00) & 0.29 & 0.26(1.10) & 0.29 & \textbf{0.26(1.10)} \\
& \textbf{DeiT\textsubscript{small}} & 0.33 & 0.28(1.10) & 0.28 & 0.28(1.00) & 0.33 & 0.26(1.20) & \textbf{0.26} & \textbf{0.26(1.00)} & 0.29 & 0.29(1.00) & 0.27 & 0.27(1.00) & 0.33 & 0.30(1.40) \\
\rowcolor{yellow!20}
& \textbf{DeiT\textsubscript{small}+CalAttn(Ours)} & 0.33 & 0.27(1.20) & 0.36 & 0.36(1.10) & 0.32 & \textbf{0.26(1.20)} & \textbf{0.28} & 0.28(1.00) & 0.29 & 0.29(1.00) & 0.28 &\textbf{ 0.26(1.10)} & 0.32 & 0.29(1.10) \\
& \textbf{Swin\textsubscript{small}} & 0.28 & 0.28(1.00) & 0.35 & 0.35(1.00) & \textbf{0.27} & \textbf{0.27(0.90)} & 0.33 & 0.30(0.80) & 0.33 & 0.29(0.80) & 0.33 & 0.31(0.80) & 0.30 & 0.29(1.20) \\
\rowcolor{yellow!20}
& \textbf{Swin\textsubscript{small}+CalAttn(Ours)} & 0.28 & 0.28(1.00) & 0.41 & 0.40(1.10) & \textbf{0.27 }& \textbf{0.27(1.00)} & 0.30 & 0.28(0.90) & 0.28 & \textbf{0.27(0.90)} & 0.27 & 0.28(0.90) & \textbf{0.27 }& \textbf{0.27(1.00)} \\

 \midrule
 
\multirow{10}{*}{CIFAR-10} & ResNet-50 & 0.87 & 0.37(2.50) & 0.46 & 0.39(1.10) & 0.97 & 0.55(2.60) & 0.80 & 0.54(0.90) & 0.41 & 0.41(1.00) & 0.45 & 0.45(1.00) & 1.18 & 0.58(1.90)\\ 
 & ResNet-110 & 1.00 & 0.54(2.60) & 0.55 & 0.46(1.20) & 1.08 & 0.60(2.80) & 0.75 & 0.50(0.90) & 0.48 & 0.46(1.10) & 0.46 & 0.52(1.10) & 1.31 & 0.68(1.90) \\ 
 & Wide-ResNet-26-10 & 0.68 & 0.34(2.20) & \textbf{0.37} & 0.37(1.00) & 0.77 & 0.41(2.20) & 0.95 & 0.37(0.80)  & 0.44 & 0.34(0.90) & 0.82 & \textbf{0.33(0.80)} & 1.24 & 1.12(1.50)\\ 
 & DenseNet-121 & 0.98 & 0.54(2.40) & 0.43 & 0.43(1.00) & 1.02 & 0.53(2.40) & 0.75 & 0.48(0.90) & 0.43 & 0.41(0.90) & 0.40 & 0.41(0.90) & 1.17 & 0.58(1.60)\\ 
& \textbf{ViT\textsubscript{224}} & 1.68 & 0.78(1.30) & 1.15 & 1.15(1.00) & 0.97 & \textbf{0.70(1.10)} & \textbf{0.73} & 0.73(1.00) & 1.76 & 1.32(0.70) & 0.96 & 0.89(0.90) & 1.44 & 1.32(1.40) \\
\rowcolor{yellow!20}
& \textbf{ViT\textsubscript{224}+CalAttn(Ours)} & 1.33 & 0.85(1.20) & 1.45 & 1.29(1.10) & 1.42 & 0.99(1.20) & 1.25 & 1.05(0.90) & 2.09 & 1.33(0.70) & 1.34 & 1.29(0.90) & \textbf{1.13} &\textbf{ 1.10(1.10)} \\
& \textbf{DeiT\textsubscript{small}} & 1.24 & \textbf{0.74(1.20)} & 1.41 & 1.41(1.00) & 1.35 & 0.71(1.20) & \textbf{0.88} & 0.88(1.00) & 2.12 & 1.05(0.70) & 1.19 & 1.08(0.80) & 1.26 & 0.97(1.30) \\
\rowcolor{yellow!20}
& \textbf{DeiT\textsubscript{small}+CalAttn(Ours)} & 1.16 & 0.89(1.10) & 0.93 & 0.86(1.10) & 1.13 & 0.91(1.10) & 0.97 & 0.97(1.00) & 1.98 & 1.14(0.70) & 1.21 & 1.16(0.90) & \textbf{0.93} & \textbf{0.86(1.10)} \\
& \textbf{Swin\textsubscript{small}} & 0.93 & 1.05(0.90) & 1.31 & 1.31(1.00) & \textbf{0.76} & 0.92(0.90) & 1.66 & \textbf{0.88(0.80)} & 3.17 & 1.15(0.60) & 1.82 & 1.15(0.70) & 1.14 & 0.78(1.20) \\
\rowcolor{yellow!20}
& \textbf{Swin\textsubscript{small}+CalAttn(Ours)} & 0.86 & 0.86(1.00) & 0.96 & 0.96(1.00) & 0.92 & 0.92(1.00) & 1.36 & 0.82(0.80) & 2.94 & 0.89(0.60) & 1.65 & 1.08(0.80) & \textbf{0.69} & \textbf{0.69(1.00)} \\

\midrule
\multirow{5}{*}{MNIST} & \textbf{ViT\textsubscript{224}} & 1.06 & 0.91(0.90) & 11.83 & 4.11(8.70) & 6.23 & 5.26(2.10) & 5.91 & 5.78(1.50) & 5.59 & 5.84(1.40) & 5.83 & 4.73(2.30) & \textbf{1.34} &\textbf{ 0.85(0.80)} \\
\rowcolor{yellow!20}
& \textbf{ViT\textsubscript{224}+CalAttn(Ours)} & 1.24 & 1.09(0.90) & 1.26 & 1.08(0.90) & 1.25 & 0.93(0.80) & 1.84 & 0.64(0.70) & 4.92 & 1.21(0.40) & 4.03 & 1.40(0.50) & \textbf{0.43 }& \textbf{0.38(0.90)} \\
& \textbf{DeiT\textsubscript{small}} &\textbf{ 1.20 }& 1.02(0.90) & 8.37 & 5.63(3.20) & 1.63 & 1.21(0.80) & 2.27 & 1.28(0.70) & 4.58 & 1.49(0.50) & 2.95 & 2.00(0.70) & 1.32 &\textbf{ 1.11(0.80)} \\
\rowcolor{yellow!20}
& \textbf{DeiT\textsubscript{small}+CalAttn(Ours)} & 1.01 & 1.01(1.00) & 1.03 & 1.03(1.00) & 1.17 & 1.03(0.90) & 1.74 & 1.04(0.80) & 4.32 & 1.75(0.50) & 3.88 & 1.36(0.60) &\textbf{ 0.53} & \textbf{0.44(0.90)} \\
& \textbf{Swin\textsubscript{small}} & 0.38 & 0.30(1.10) & 0.53 & 0.43(0.90) & 0.71 & 0.63(0.90) & 2.21 & 1.01(0.70) & 1.60 & 0.36(0.50) & 1.29 & 0.53(0.60) & \textbf{0.30} & \textbf{0.30(1.00)} \\
\rowcolor{yellow!20}
& \textbf{Swin\textsubscript{small}+CalAttn(Ours)} & 0.96 & 0.46(1.10) & 0.42 & 0.37(0.90) & 0.68 & 0.55(0.90) & 2.12 & 0.82(0.70)  & 2.51 & 0.49(0.50) & 1.96 & 0.44(0.50) &\textbf{ 0.32} &\textbf{ 0.27(0.90)} \\
\midrule
\multirow{5}{*}{Tiny-ImageNet} 
& \textbf{ViT\textsubscript{224}} & 0.14 & 0.12(1.10) & 0.14 & 0.12(1.10) & 0.14 & 0.11(1.20) & 0.12 &\textbf{ 0.11(1.10)} & 0.13 & 0.12(1.10) & \textbf{0.12} & 0.11(1.10) & 0.13 & 0.12(1.10) \\
\rowcolor{yellow!20}
& \textbf{ViT\textsubscript{224}+CalAttn(Ours)} & 0.33 & 0.14(1.90) & 0.18 & 0.12(1.30) & 0.29 & 0.14(1.80) & 0.11 & 0.10(1.10) & 0.11 & 0.10(1.00) & 0.24 & 0.14(1.40) & \textbf{0.10} &\textbf{ 0.10(1.00)} \\
& \textbf{DeiT\textsubscript{small}} & 0.12 & 0.11(1.10) & 0.11 & 0.10(1.10) & 0.12 & 0.10(1.10) & 0.11 & 0.11(1.00) & 0.11 & 0.11(1.00) & 0.11 & 0.11(1.00) & 0.12 & 0.11(1.10)\\
\rowcolor{yellow!20}
& \textbf{DeiT\textsubscript{small}+CalAttn(Ours)} & 0.23 & 0.14(1.40) &\textbf{ 0.07} & \textbf{0.06(1.20)} & 0.22 & 0.13(1.50) & 0.12 & 0.11(1.10) & 0.21 & 0.13(1.30) & 0.19 & 0.14(1.30) & 0.11 & 0.10(1.10)  \\
& \textbf{Swin\textsubscript{small}} & 0.13 & 0.15(0.90) & \textbf{0.06} & \textbf{0.07(0.90)} & 0.12 & 0.12(1.00) & 0.13 & 0.13(0.90) & 0.13 & 0.14(0.90) & 0.13 & 0.13(0.90) &  0.13 & 0.13(1.00)\\
\rowcolor{yellow!20}
& \textbf{Swin\textsubscript{small}+CalAttn(Ours)} & 0.16 & 0.14(1.10) & 0.11 & 0.11(1.10) & 0.12 & 0.12(1.00) & 0.12 & 0.12(1.00) & 0.14 & 0.14(1.00) & 0.14 & 0.14(1.00) &\textbf{0.06} & \textbf{0.07(0.90)}  \\
\midrule
\bottomrule
\end{tabular}
}
\end{table*}

\end{document}